\documentclass[a4paper]{article}

\usepackage[english]{babel}
\usepackage[utf8]{inputenc}
\usepackage{amsmath}
\usepackage{graphicx}
\usepackage[colorinlistoftodos]{todonotes}
\usepackage{geometry}
\usepackage{subfigure}
\usepackage{caption}
\usepackage{algorithm}
\usepackage{algpseudocode}
\usepackage{pifont}
\usepackage{multirow}
\usepackage{amsfonts}
\usepackage{lineno}
\usepackage[percent]{overpic}
\usepackage{todonotes} % para agregar comentarios en el pdf, para correcciones de los revisores
\usepackage{soul} % para resaltar el texto que se quiere remarcar
\definecolor{myc}{HTML}{3A08C8}
\usepackage[toc,page]{appendix}

\renewcommand{\vec}[1]{\boldsymbol{\mathbf{#1}}}

\title{A multi-class structured dictionary learning method using discriminant atom selection}
\author{R.E. Rolón$^{a,d}$, L.E . Di Persia$^a$, R.D. Spies$^b$ and H.L. Rufiner$^{a,c}$}

\date{\today}

\begin{document}

\maketitle

\begin{center}
$^a$ Instituto de Investigaci\'{o}n en Se\~{n}ales, Sistemas e Inteligencia Computacional, sinc($i$), FICH, UNL, CONICET, Santa Fe, Argentina
	
$^b$ Instituto de Matem\'{a}tica Aplicada del Litoral, IMAL, FIQ, UNL, CONICET, Santa Fe, Argentina

$^c$ Laboratorio de Cibern\'{e}tica, Fac. de Ing., Univ. Nacional de Entre R\'{i}os, Argentina
	
$^d$ Facultad Regional Paraná, Universidad Tecnológica Nacional, Entre Ríos, Argentina
	
\end{center}

\begin{abstract}
In the last decade, traditional dictionary learning methods have been successfully applied to various pattern classification tasks. Although these methods produce sparse representations of signals which are robust against distortions and missing data, such representations quite often turn out to be unsuitable if the final objective is signal classification. In order to overcome or at least to attenuate such a weakness, several new methods which incorporate discriminative information into sparse-inducing models have emerged in recent years. In particular, methods for discriminative dictionary learning have shown to be more accurate (in terms of signal classification) than the traditional ones, which are only focused on minimizing the total representation error. In this work, we present both a novel multi-class discriminative measure and an innovative dictionary learning method. For a given dictionary, this new measure, which takes into account not only when a particular atom is used for representing signals coming from a certain class and the magnitude of its corresponding representation coefficient, but also the effect that such an atom has in the total representation error, is capable of efficiently quantifying the degree of discriminability of each one of the atoms. On the other hand, the new dictionary construction method yields dictionaries which are highly suitable for multi-class classification tasks. Our method was tested with a widely used database for handwritten digit recognition and compared with three state-of-the-art classification methods. The results show that our method significantly outperforms the other three achieving good recognition rates and additionally, reducing the computational cost of the classifier.
\end{abstract}

%The original motivation for this approach was the strong correlation observed in a previous work \cite{rolon_discriminative_2017}, where the activations of the atoms in a dictionary surprisingly have important discriminative information regarding class membership in the context of a binary classification problem
\section{Introduction}
Sparse representation of signals is considered a very powerful signal processing technique which has drawn massive interest in recent years mainly due to its success in solving a wide variety of problems in different fields such as biomedical signal processing \cite{rolon_discriminative_2017,peterson_generalized_2017}, computer vision \cite{li_learning_2014} and image analysis \cite{mairal_discriminative_2008}, including image denoising \cite{elad_image_2006}, color image restoration \cite{mairal_sparse_2008} and image classification \cite{zhang_discriminative_2010}. Roughly speaking, the problem of sparse representation consists of obtaining approximations of the involved signals in terms of linear combinations of only a few prescribed very simple characteristic signals taken from a large set \cite{lewicki_probabilistic_1999,aharon_ksvd_2006}. Besides providing a robust framework against distortions, missing data and noise, sparse representation of signals has many other advantages such as super resolution and dimensionality reduction \cite{yang_image_2010}.

A sparse representation problem (SRP) is usually divided into two sub-problems: an inference problem and a learning problem. The first one, which is often called ``sparse coding'',  consists of computing a representation vector satisfying a particular sparsity constraint given a predefined dictionary. The second one, which involves solving a more complex problem, consists of finding an ``optimal'', in certain sense, dictionary for representing a given set of training signals. It is important to point out however, that most formulations of SRPs only focus on minimizing a prescribed total representation error and they do not take into account any a-priori discriminative information which could significantly improve the performance in the case of multi-object classification problems.

The first data-driven dictionary learning algorithms were originally developed almost two decades ago \cite{lewicki_probabilistic_1999,lewicki_learning_2000,engan_method_1999}. Some of them have their roots in probabilistic frameworks by considering the observed data as realizations of certain random variables \cite{lewicki_probabilistic_1999,lewicki_learning_2000}. In \cite{lewicki_learning_2000} for example, the authors developed an algorithm for finding a redundant dictionary maximizing the likelihood function of the probability distribution of the data. In that work, an analytic expression for the likelihood function was derived by approximating the posterior distribution by Gaussian functions. On the other hand, an iterative approach for dictionary learning, known as the ``Method for Optimal Directions'' (MOD), was presented in \cite{engan_method_1999}. The sparse coding stage of this method makes use of a greedy algorithm called ``Orthogonal Matching Pursuit'' (OMP) \cite{tropp_signal_2007} followed by a simple dictionary updating rule.

A new iterative algorithm was proposed by Aharon \textit{et al.} in \cite{aharon_ksvd_2006}. This new approach, called ``K Singular Value Decomposition'' (KSVD), consists mainly of two stages: a sparse coding stage and a dictionary learning stage. The OMP algorithm is used in the sparse coding stage, which is followed by a dictionary updating step where the atoms are updated one at a time and the representation coefficients are allowed to change in order to minimize the total representation error.

In the last decade, the interest in developing algorithms based on sparse representation of signals for pattern recognition purposes has notably increased \cite{zhang_discriminative_2010,pham_joint_2008,jiang_label_2013}. This is so because a large number of authors have proposed new supervised approaches for pattern recognition using sparse representations of signals. For instance, a discriminative version of the standard KSVD method applied to face recognition was presented by Zhang Q. \textit{et al.} \cite{zhang_discriminative_2010}. In that work, the authors included a discriminative term into the objective function of the standard KSVD algorithm. Results have shown that this modification constitutes an appropriate way to learn dictionaries which satisfy both criteria: low reconstruction error and high recognition rates. Also, Pham D. \textit{et al.} \cite{pham_joint_2008} proposed an iterative method that simultaneously optimizes a dictionary and a linear classifier. The authors successfully used the method in an image categorization problem. More recently, a novel approach called ``Label Consistent KSVD'' (LC-KSVD) for dictionary learning was proposed in \cite{jiang_label_2013}. In that work a discriminative sparse representation and a single predictive linear classifier were efficiently integrated into the objective function.

However, besides supervised dictionary learning methods, many other new alternative options were presented \cite{mairal_discriminative_2008,zhang_learning_2009,yang_unifying_2008}. These new alternatives are mainly based on the pursuit of discriminability of sparse representations through the development of ``structured'' or, more precisely, category-specific dictionary methods. In \cite{mairal_discriminative_2008}, a method for learning multiple dictionaries that uses the reconstruction
errors yielded by these dictionaries on image patches to derive
a pixel-wise classification. This algorithm has proved to be robust specially for local image classification tasks. A method for learning multiple non-redundant dictionaries for complex object categorization was proposed in \cite{zhang_learning_2009}. This method was assessed on both visual object categorization and document classification image-related problems yielding competitive performances. In \cite{yang_unifying_2008}, a method that simultaneously optimizes both a structured dictionary (category-specific visual words for each feature) and a classifier was introduced. This method yielded good recognition rates showing a significant improvement over state-of-the-art object classification methods. A new method for structured dictionary learning was recently proposed by Sun \textit{et al.} \cite{sun_sparse_2018}. In that work, the learned dictionary was decomposed into class-specific sub-dictionaries for the classification that is conducted measuring the minimum reconstruction error among all the classes. The method was tested using both the synthetic data and the real-world data showing good performances.

In this work we propose a novel multi-class discriminative measure and a new dictionary learning method which yields structured dictionaries which are composed by category-specific sub-dictionaries specially constructed for multi-class classification purposes. Thus, the novelty of our approach is twofold. First, we introduce an innovative and effective multi-class discriminative measure whose main property is precisely its capability for quantifying the discriminative degree of each one of the atoms in a given dictionary. This measure takes into account not only when a particular atom is used for representing a signal coming from a certain class and the magnitude of its corresponding representation coefficient, but also the effect that such an atom has in the total representation error. Secondly, this work presents a novel method for discriminative structured dictionary learning which yields a dictionary increasing the classifier recognition rate.

The organization of this article is as follows. A brief review of sparse representation of signals is presented in Section \ref{sec: sparse representation}. In Section \ref{sec: the method}, we make a description of the database used in the experiments and we propose both a new discriminative measure and a structured dictionary learning method. Section \ref{sec: experiments} details all the experiments, while results and discussion are presented in Section \ref{sec: results and discussion}. Finally, concluding comments and future works are presented in Section \ref{sec: conclusions}.
%---------
\section{Sparse representation of signals}\label{sec: sparse representation}
Sparse representation is a signal processing technique that seeks the sparsest representation of all the signals in a given set in terms of linear combinations of certain basic waveforms. The sparse representation problem can be separated into two sub-problems. Namely the so-called sparse coding problem and the dictionary learning problem. We shall now proceed to describe in detail each one of these sub-problems. For that, let $\vec{x}\in\mathbb{R}^N$ be a discrete signal and let $\Phi\in\mathbb{R}^{N\times M}$ (generally with $M \geq N$) be a dictionary whose columns $\vec{\phi}_j\in\mathbb{R}^N$ are atoms that we want to use for obtaining representations of $\vec{x}$ of the form $\vec{x}=\Phi \vec{a}$. Here, and in the sequel, we shall refer to the vector $\vec{a}=[a_1 \; a_2 \; \cdots \; a_M]^T \in\mathbb{R}^M$ as a ``representation'' of $\vec{x}$. Sparsity consists essentially of obtaining a representation with as few non-zero elements as possible. A way of obtaining such representations consists of solving the following problem: 
\begin{equation*}
\left(P_0\right): \;\;\; \underset{\vec{a} \in \mathbb{R}^M}{\textrm{min}} \; {||\vec{a}||}_{0} \; \textrm{subject to} \; \vec{x} = \Phi \vec{a},
\label{p_zero}
\end{equation*}
\noindent where $||\vec{a}||_0$ denotes the $\ell_0$ pseudo-norm, defined as the number of non-zero elements of $\vec{a}$. It turns out that imposing an exact representation of $\vec{x}$ is a too restrictive constraint, which makes $\left(P_0\right)$ an NP hard problem \cite[\S 1.8]{elad_sparse_2010}, yielding the approach highly unsuitable for most practical applications.

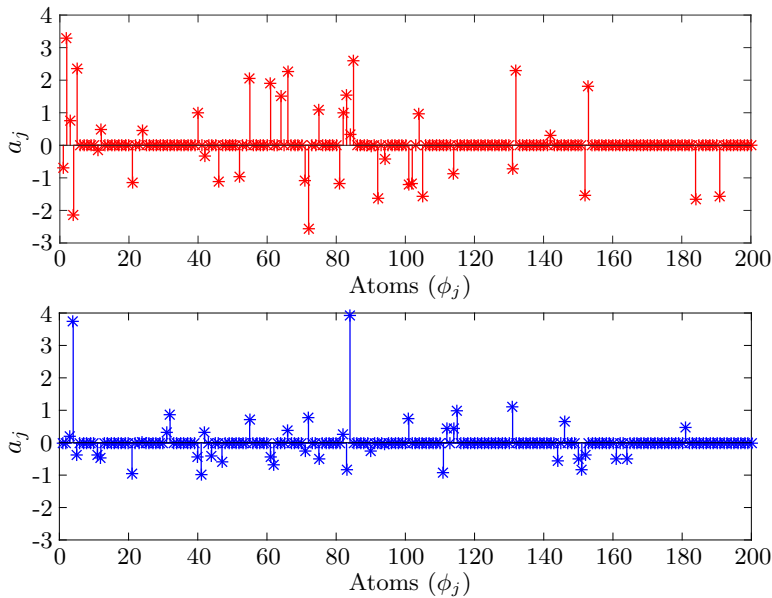
\begin{figure}[!h]
	\centering
	\begin{overpic}[width=4in]{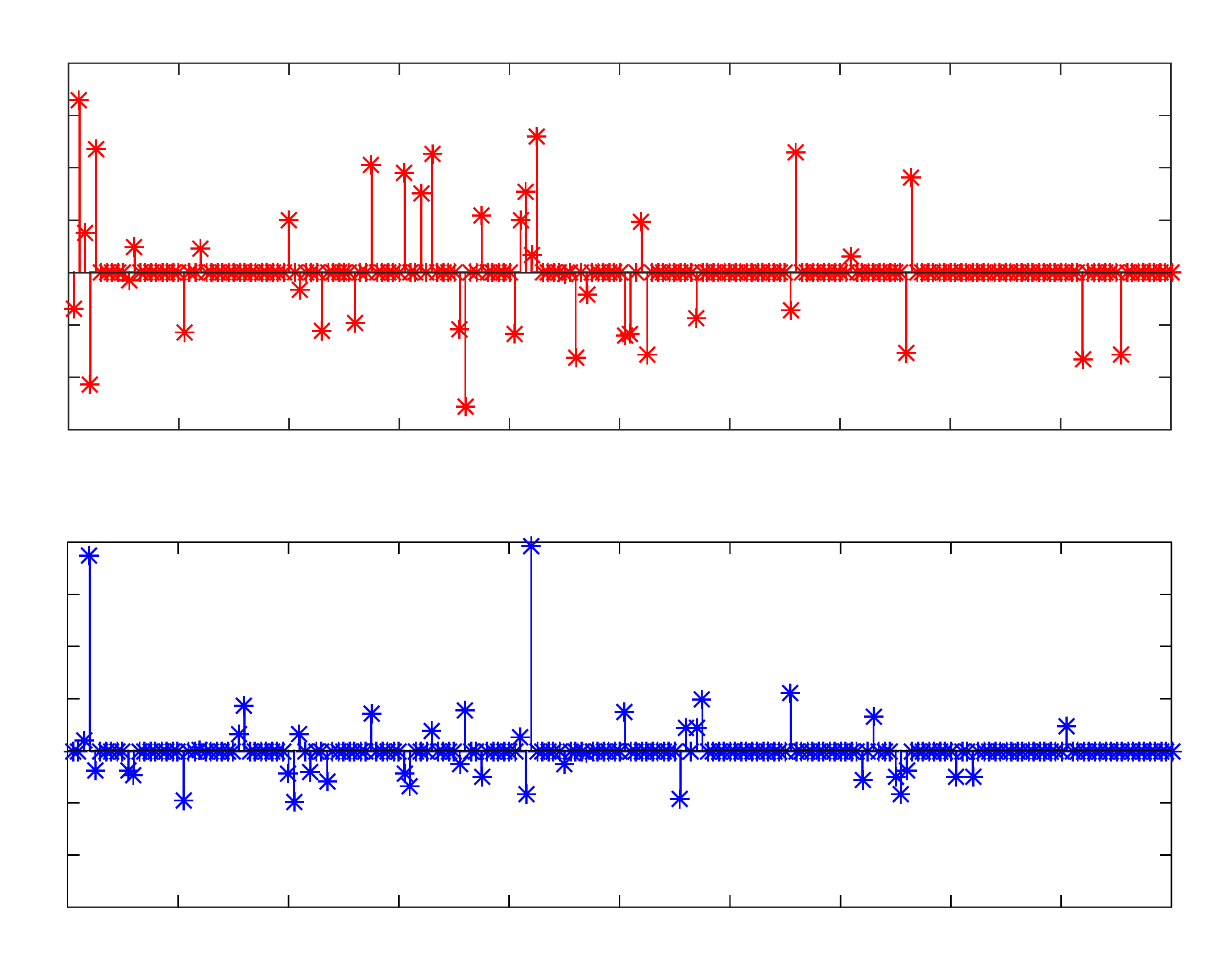}
		%\begin{overpic}[width=4in,grid,tics=5]{images/sparse_code_class_zero_one.pdf}
		\put(-1,15.8){\rotatebox{90}{$a_j$}}
		\put(43,-2){\small Atoms $\left(\phi_j\right)$}
		\put(-1,55){\rotatebox{90}{$a_j$}}
		\put(43,36.8){\small Atoms $\left(\phi_j\right)$}
		\put(4.8,1.5){\small 0}
		\put(13,1.5){\small 20}
		\put(22,1.5){\small 40}
		\put(31,1.5){\small 60}
		\put(40,1.5){\small 80}
		\put(48,1.5){\small 100}
		\put(57,1.5){\small 120}
		\put(66,1.5){\small 140}
		\put(75,1.5){\small 160}
		\put(84,1.5){\small 180}
		\put(93,1.5){\small 200}
		\put(4.8,40.4){\small 0}
		\put(13,40.4){\small 20}
		\put(22,40.4){\small 40}
		\put(31,40.4){\small 60}
		\put(40,40.4){\small 80}
		\put(48,40.4){\small 100}
		\put(57,40.4){\small 120}
		\put(66,40.4){\small 140}
		\put(75,40.4){\small 160}
		\put(84,40.4){\small 180}
		\put(93,40.4){\small 200}
		\put(2,3.5){\small -3}
		\put(2,7.7){\small -2}
		\put(2,12){\small -1}
		\put(3,16.2){\small 0}
		\put(3,20.5){\small 1}
		\put(3,24.7){\small 2}
		\put(3,29){\small 3}
		\put(3,33){\small 4}
		\put(2,42.4){\small -3}
		\put(2,46.5){\small -2}
		\put(2,50.8){\small -1}
		\put(3,55){\small 0}
		\put(3,59.2){\small 1}
		\put(3,63.6){\small 2}
		\put(3,67.7){\small 3}
		\put(3,72){\small 4}
	\end{overpic}
	\caption{Example of two representation vectors of handwritten digits of two different classes obtained with the OMP algorithm.}
	\label{fig: sparse_codes}
\end{figure}

Hence, the exact representation requirement $\vec{x} = \Phi \vec{a}$ is often relaxed by allowing small representation errors and imposing an upper bound on the $\ell_0$ pseudo-norm of the representations. Thus, a small error representation tolerant version of $\left(P_0\right)$ is defined as follows:
\begin{equation*}
\left(P_0^q \right): \;\;\; \underset{\vec{a} \in \mathbb{R}^M}{\textrm{min}} \; ||\vec{x} - \Phi \vec{a}||_2^2 \; \textrm{subject to} \; ||\vec{a}||_0 \leq q,
\label{p_two}
\end{equation*}
\noindent where $q$ is a prescribed positive integer parameter. This formulation considers the presence of possible additive noise terms. In other words, it assumes that the signal $\vec{x}$ can be represented in the form $\vec{x}=\Phi \vec{a} + \vec{e}$, where $\vec{e} \in \mathbb{R}^{N}$ is a small energy noise term. Thus, this approach is more appropriate in a wide variety of real applications (such as biomedical signal and image processing) where the captured raw signals are always contaminated by noise. Several greedy strategies have been proposed for solving problem $\left(P_0^q \right)$ \cite{mallat_matching_1993,tropp_signal_2007}. Among them, the OMP algorithm is perhaps the most commonly used strategy. This greedy algorithm ensures convergence to the projection of $\vec{x}$ into the span of atoms in a given dictionary, in no more than $q$ iterations. It is important to note that the representation vector $\vec{a}$ has no more than $q \ll M$ non-zero entries. \figurename{ \ref{fig: sparse_codes}} shows an example of the representation vectors obtained with this $\left(P_0^q \right)$ approach for two images of different classes coming from a widely used database which we shall describe in detail in Section \ref{sec: the method}. Note that most coefficients are strictly equal to zero.

Although pre-constructed dictionaries, such as the well known wavelet packets \cite{coifman_signal_1994}, typically lead to fast sparse coding, they are almost always highly restricted to certain classes of signals. Hence, due to their lack of generalization, new approaches introducing data-driven dictionary learning techniques have emerged. A dictionary learning problem associated to the data: $q$, $M$, $N \in \mathbb{N}$, $M \geq N$ and a collection of $n$ signals in $\mathbb{R}^N$, $\vec{x}_1,\cdots,\vec{x}_n$, can be formally written as: 
\begin{equation*}
\left(DL \right): \;\;\; \underset{\underset{\vec{a}_i \in \mathbb{R}^M,||\vec{a}_i||_0\leq q,1\leq i \leq n.}{\Phi \in \mathbb{R}^{N \times M}}}{\textrm{min}} \; \sum_{i=1}^n||\vec{x}_i-\Phi \vec{a}_i||_2^2
\label{p_dict_learning}
\end{equation*}
\noindent The solution of this problem yields on one hand a dictionary $\Phi$ and, on the other hand, representations for all the signals in terms of that dictionary complying with the sparsity constraint for each one of the ``$n$'' involved signals $\vec{x}_1,\cdots,\vec{x}_n$. It is important to point out that in such a process, the total representation error is minimized.

Although data-driven dictionary learning algorithms produce sparse representations of signals which are robust against distortions and missing data, such representations quite often turn out to be unsatisfactory if the final objective is signal classification. This is mainly due to the fact that those algorithms do not take into account prior information concerning class membership. To overcome this flaw, several alternative approaches producing sparse representations in terms of a unique (and shallow) dictionary for signal classification were presented \cite{zhang_discriminative_2010,pham_joint_2008,jiang_label_2013}. 
A different approach is the construction of structured dictionaries composed by sub-dictionaries whose atoms are discriminative, in certain sense, for each one of the classes, i.e. each sub-dictionary has a group of atoms that are discriminative only for a particular class. The use of structured dictionaries could be useful for reducing the features dimension, avoiding over-fitting and optimizing the performance of a classifier, among others. In recent years, there has been increasing interest in developing algorithms whose main purpose is to obtain ``optimal'' sub-dictionaries to be used for signal classification \cite{rolon_discriminative_2017,rao_clustering_2012,chen_learn_2016}. In \cite{rao_clustering_2012}, a method called ``Clustering based Online Learning of Dictionaries'' (COLD) was presented. This algorithm makes use of the mean shift clustering procedure \cite{comaniciu_mean_2002} to identify modes in the distribution of the atoms and hence obtain a dictionary of minimal size. Recently, Chen \textit{et al.} \cite{chen_learn_2016} introduced a dictionary learning method for image and video editing tasks. In that work, the problem of seeking an optimal dictionary is solved by using a symmetric version of the ``Kullback-Leibler Divergence'' (KLD) \cite{jeffreys_invariant_1946}. This divergence has been successfully used for detecting redundant atoms in a given dictionary. Our proposal consists of defining and using a new discriminative measure for selecting the most discriminative atoms for each one of the classes and use them for building a new structured dictionary.

\section{Materials and methods}\label{sec: the method}
In this section we make a brief description of the database used in the experiments. Additionally, we describe in detail both the new proposed multi-class discriminative measure and the novel structured dictionary learning method.
%---

\subsection{Database}\label{sec: database}
One of the most popular databases used to assess Computer Vision and Pattern Recognition methods is the ``Modified NIST'' (MNIST) database \cite{lecun_comparison_1995}. This database has been widely used for assessing new methods including Deep Learning techniques \cite{kim_deep_2015}, Extreme Learning Machines \cite{chazal_comparison_2015} and a many types of neural networks \cite{lecun_gradient-based_1998}, among others. The MNIST database contains a total of 70,000 normalized and centered gray-scale images of handwritten digits ranging from 0 (zero) to 9 (nine), each one of size $28 \times 28$ (leading to a feature vector of length 784). \textcolor{black}{Also, the number of images per class varies from 5,421 to 6,742, corresponding to classes 5 and 1, respectively.} Additionally, this database provides information about standard partitions used for training (60,000) and testing (10,000).

Although each one of all original (raw) images coming from the MNIST database can be represented as a single column vector consisting of 784 elements (features), it becomes necessary to reduce its dimensionality for practical reasons. In this work, the image dimension reduction process is carried out by using the well known bi-cubic interpolation method \cite{russell_polynomial_1995} which is not only accurate, but also smooth and computationally efficient. This method was used for obtaining new (reduced) images each one of size $16 \times 16$ leading to feature vectors of length 256.

\subsection{A new discriminative measure} \label{multi_class_disc_inf}
Discriminative dictionaries can be thought of as a collection of atoms specially learned for signal classification. These dictionaries not only produce accurate representations of the training signals (in terms of their waveforms) coming from different classes, but they also render their representations easy to distinguish by a suitable classifier. However, the problem of finding a discriminative dictionary is computationally very costly. A way to overcome the computational complexities entailed by such a problem consists of defining an appropriate discriminative value functional that independently evaluates each one of the atoms in a given dictionary. This simplification is based on the assumption that each atom in the dictionary is used to model specific characteristics that are not modeled by any one of the other atoms. Thus, the discriminative information provided by a particular atom is different from the information contributed by all the other atoms.

In a previous work \cite{rolon_discriminative_2017}, we presented a simple approach for quantifying the discriminative degree of the atoms of a given dictionary $\Phi$ in the context of a binary classification problem. The approach essentially consists of counting the number of times that a particular atom is used, i.e. it becomes ``active'' for representing signals belonging to each one of both classes $\ell=1$ and $\ell=2$. As a result of this counting process, an activation frequency ($\eta$) for each atom given the class, is considered. To quantify the discriminative degree of the $j^{\textrm{th}}$-atom ($\phi_j$, the $j^{\textrm{th}}$-column of $\Phi$), the absolute difference of activation frequencies of that atom for classes $\ell=1$ and $\ell=2$ ($|\eta_{1}^j-\eta_{2}^j|$) is computed. This value will be large if (an only if) the atom $\phi_j$ is much more frequently used for representing signals in one of the two classes and, in that case, it can be thought of as a quantifier of the capability of $\phi_j$ to supply important discriminative information regarding class membership. The use of this discriminative quantifier gave rise to a method called Most Discriminative Column Selection (MDCS) for discriminative sub-dictionary construction \cite{rolon_discriminative_2017}. The MDCS method has shown to be robust for efficiently extracting meaningful features from segments of pulse oximetry signals for detecting apnea-hypopnea events.

In this work we propose an extension of the measure described above to multi-class classification problems. This extension consists of defining and using a new multi-objective function aimed at quantifying the discriminative degree of each one of the atoms in a given dictionary. This function will be defined as a convex combination of three discriminative terms, all based on the affine sparse representations of the data. In what follows, a detailed description of each one of such terms as well as a formal definition of the function are presented.
%--
\subsubsection{Activation frequency measure}
Conditional activation frequencies provide a reasonable starting point for determining the discriminative degree of individual atoms in a given dictionary. For this reason, our approach begins by computing the activation frequency $\eta_{\ell}^j$ of $\phi_j$ given the class $\ell$, for $\ell=1,2,\cdots,k$. Moreover, the conditional activation probability of $\phi_j$ given (that a signal $\vec{x}$ belongs to) class $\ell$ is defined as $p_\ell^j \doteq P(a_j \not = 0|\vec{x} \in \ell)$. Given a set of $n_\ell$ signals belonging to class $\ell$, this conditional probability can be approximated by the quotient $\eta_\ell^j/n_\ell$. Note that if the problem is balanced, i.e. if the number of available signals belonging to each one of the $k$ classes is the same, say $\hat{n}$, then $\eta_\ell^j \propto p_\ell^j$, more precisely $\eta_\ell^j=k \hat{n} p_\ell^j$, for all $\ell$ and $j$. In this work, the problem of quantifying the discriminability of each atom is tackled by analyzing their individual contributions to the signal classification process. More specifically, a particular atom $\phi_j$ is considered as having important discriminative information for class $\ell$ signals if $p_\ell^j>p_m^j$, for all $m \not = \ell$. Hence, if $\phi_j$ is discriminative for class $\ell$, the activation of the representation coefficient $a_j$ will be strongly associated to class $\ell$ membership. Since the performance of a classifier highly depends on the discriminability of their inputs, it is reasonable to think that using the representation coefficients $a_1$, $a_2$, $\cdots$, $a_M$ as inputs of a classifier, for atoms selected using that criterion, could result in good recognition rates.

For a given $j$, $1 \leq j \leq M$, we shall denote by $\ell_j^+$ the class that maximizes all conditional activation probabilities $p_\ell^j$, for all $\ell=1,2,\cdots,k$, i.e. such that
\begin{equation}\label{ell_sum}
p^j_{\ell_j^+} = \underset{1\leq \ell \leq k}{\textrm{max}} \; p_\ell^j.
\end{equation}
\noindent In the (unlikely) case that there is more than one value of $\ell$ maximizing $p_\ell^j$, $\ell_j^+$ is defined by randomly choosing one of them, for instance the smallest one (note that the order of the classes is completely irrelevant). Similarly, for a fixed $j$, $1 \leq j \leq M$, $\ell_j^*$ is defined as the class leading to the second largest conditional activation probability, i.e. such that
\begin{equation}\label{ell_star}
p^j_{\ell_j^*} = \underset{\underset{\ell \not = \ell_j^+}{1\leq \ell \leq k}}{\textrm{max}} \; p_\ell^j.% \;\textrm{and} \; \eta_m^j<\eta_\ell^j.
\end{equation}
\noindent Here again if there is more than one value of $\ell_j^*$ satisfying (\ref{ell_star}), then $\ell_j^*$ is chosen randomly as any one of them.

Next we define the function $m_{af}: \; \{1,2,\cdots,M\} \rightarrow \mathbb{R}_0^+$ by
\begin{equation}
m_{af}(j) \doteq \frac{p_{\ell_j^+}^j-p_{\ell_j^*}^j}{p_{\ell_j^+}^j},
\end{equation}
\noindent we shall refer to $m_{af}(\cdot)$ as the ``activation frequency measure''.

Note that $0 \leq m_{af}(\cdot) \leq 1$. The atom $\phi_j$ is said to be discriminative (for class $\ell_j^+$) if and only if $m_{af}(j)>0$. Clearly, within this setting, if an atom $\phi_j$ is discriminative, it will be so only for the class $\ell_j^+$, otherwise it will be discriminative for none of them. Moreover, the value of $m_{af}(j)$ can be thought of as a ``measure'' of the degree of discriminability of the atom $\phi_j$ (for the corresponding class $\ell_j^+$), based solely on the activation frequency information.

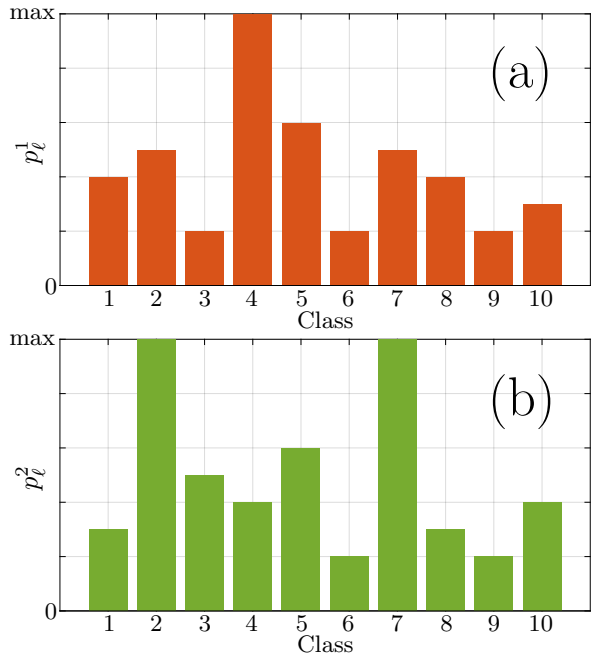
\begin{figure}[h!]
	\centering
	\begin{overpic}[width=3in]{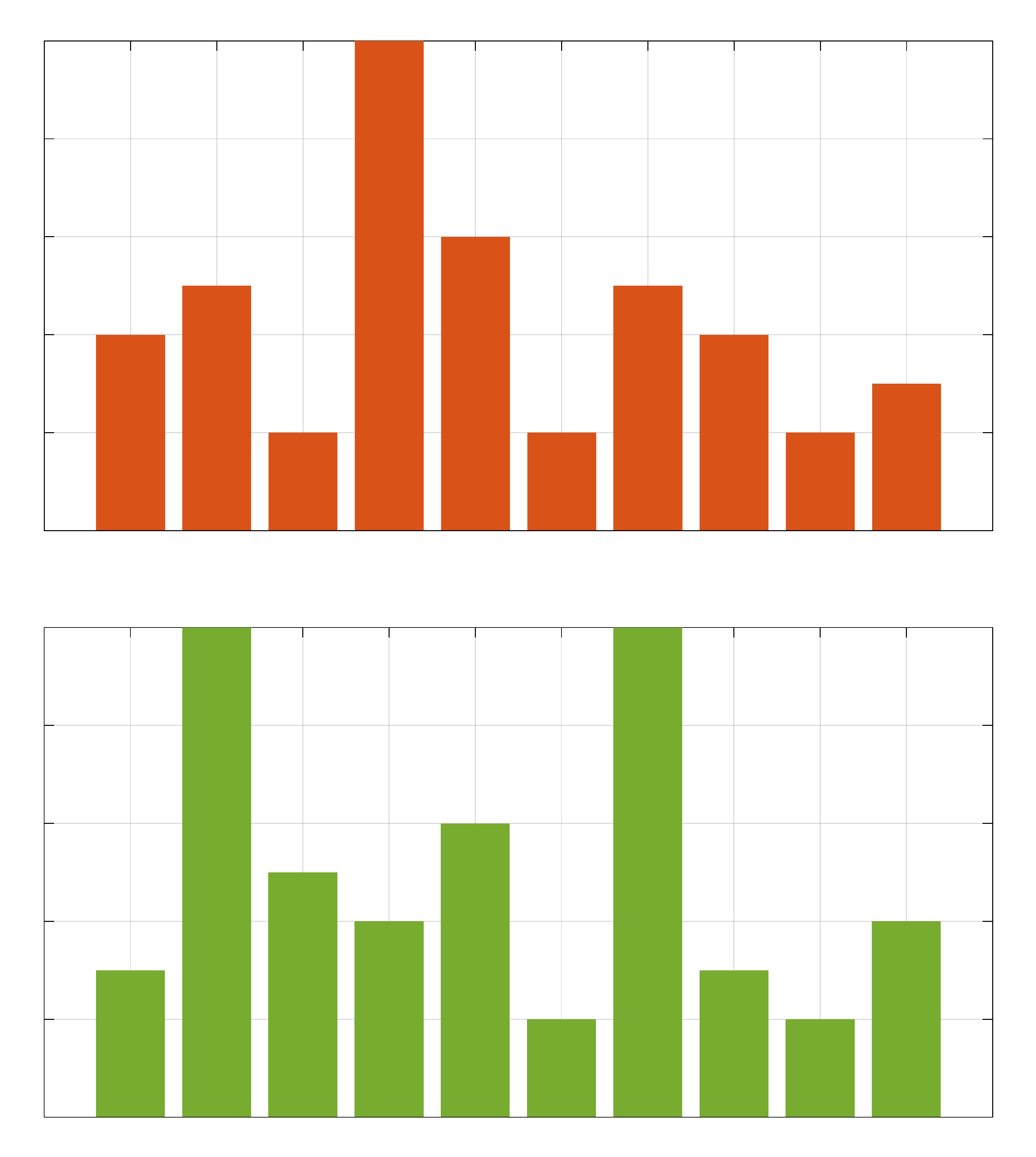}
		%\begin{overpic}[width=3in,grid,tics=5]{images/discriminative_measure_examples.pdf}
		\put(10.5,51){\small 1}
		\put(17.8,51){\small 2}
		\put(25.3,51){\small 3}
		\put(32.5,51){\small 4}
		\put(40.3,51){\small 5}
		\put(47.5,51){\small 6}
		\put(55,51){\small 7}
		\put(62.7,51){\small 8}
		\put(70,51){\small 9}
		\put(76.3,51){\small 10}
		\put(10.5,0.3){\small 1}
		\put(17.8,0.3){\small 2}
		\put(25.3,0.3){\small 3}
		\put(32.5,0.3){\small 4}
		\put(40.3,0.3){\small 5}
		\put(47.5,0.3){\small 6}
		\put(55,0.3){\small 7}
		\put(62.7,0.3){\small 8}
		\put(70,0.3){\small 9}
		\put(76.3,0.3){\small 10}
		\put(1.5,2.2){\small 0}
		\put(1.5,52.8){\small 0}
		\put(-4,44.9){\small max}
		\put(-4,95.6){\small max}
		\put(-3,73){\rotatebox{90}{$p^1_\ell$}}		
		\put(-3,22.2){\rotatebox{90}{$p^2_\ell$}}
		\put(40.5,47.5){\small Class}
		\put(40.5,-3){\small Class}
		\put(70,85){\huge (a)}
		\put(70,33.9){\huge (b)}
	\end{overpic}
	\caption{Vertical bars representing conditional activation probabilities for a class-specific discriminative atom $\phi_1$ (top) and for a non-discriminative atom $\phi_2$ (bottom) as proposed in this work.}
	\label{label: two_examples_disc_non_disc_atoms}
\end{figure}

\figurename{ \ref{label: two_examples_disc_non_disc_atoms}} shows graphic representations of two examples of conditional activation probabilities $p^1_\ell$ and $p^2_\ell$, for $\ell = 1,2,\cdots,10$, associated to atoms $\phi_1$ (top) and $\phi_2$ (bottom), respectively. The vertical bars represent the value of each conditional activation probability $p_\ell^1$ (top) and $p_\ell^2$ (bottom), for $\ell = 1,2,\cdots,10$. Clearly, for the top case (atom $\phi_1$) $\ell_1^+=4$ and $\ell_1^*=5$, $m_{af}(1)>0$ and therefore the atom $\phi_1$ is considered to be discriminative (for class 4). For the bottom case (atom $\phi_2$) $\ell_2^+=2$ and $\ell_2^*=7$ (although these values could be interchanged), but since $p_2^2=p_7^2$, one has $m_{af}(2)=0$ implying that $\phi_2$ is not discriminative for class $\ell_2^+$, and therefore is not discriminative for any one of the classes.
%---
%---
\subsubsection{Coefficient magnitude measure}
On one hand, the sparse representation of signals provides valuable information regarding the activation of atoms and, on the other hand, it can highlight important characteristics or features contained in particular event related waveforms of signals or images such as brightness variations in images and slight changes in biomedical signals, to name but a few. With the above observation in mind, we proceed now to define a second measure that takes into account the magnitude of the representation coefficients. For that, given an atom $\phi_j$, let $\ell_j^+$ and $\ell_j^*$ be the classes as defined in (\ref{ell_sum}) and (\ref{ell_star}), respectively, and let $\vec{A}_{\ell_j^+}$ and $\vec{A}_{\ell_j^*}$ the matrices which provide the sparse representations of $\vec{X}_{\ell_j^+}$ and $\vec{X}_{\ell_j^*}$, respectively, in terms of the dictionary $\Phi$, i.e. $\vec{X}_{\ell_j^+}=\Phi \vec{A}_{\ell_j^+}$ and $\vec{X}_{\ell_j^*}=\Phi \vec{A}_{\ell_j^*}$. Additionally, let $q_\ell^j$ denote the quotient $||\left[\vec{A}_\ell\right]_{j,:}||_1/n_\ell$, where $\left[\vec{A}_\ell\right]_{j,:}$ represents the $j^\textrm{th}$-row of the matrix $\vec{A}_\ell$. The coefficient magnitude measure is the function $m_{cm}: \; \{1,2,\cdots,M\} \rightarrow \mathbb{R}_0^+$ defined by
\begin{equation}
m_{cm}(j) \doteq \frac{q_{\ell_j^+}^j-q_{\ell_j^*}^j}{q_{\ell_j^+}^j}.
\end{equation}

Here again $0 \leq m_{cm}(\cdot) \leq 1$. Based on this measure, an atom $\phi_j$ is said to be discriminative (for the class $\ell_j^+$) if and only if $m_{cm}(j)>0$ and, in that case, the value of $m_{cm}(j)$ quantifies the corresponding degree of discriminability of $\phi_j$ for the class $\ell_j^+$.
\subsubsection{Representation error measure}
We now proceed to describe the third measure for quantifying the discriminative degree of each atom in a dictionary. This measure takes into account the contribution of each atom $\phi_j$ to the total representation error. Let $\vec{A}_\ell \doteq [\vec{a}_1 \; \vec{a}_2 \; \cdots \vec{a}_{n_\ell}]$ be the matrix providing the sparse representation of $\vec{X}_\ell \doteq [\vec{x}_1 \; \vec{x}_2 \; \cdots \vec{x}_{n_\ell}]$, as in the previous measure. Clearly, the contribution of the class $\ell$ to the total representation error can be written as \cite{aharon_ksvd_2006}
\begin{eqnarray}
\sum_{i=1}^{n_\ell}||\vec{x}_i-\Phi \vec{a}_i||_2^2&=&\left\|\vec{X}_\ell-\Phi \nonumber \vec{A}_\ell\right\|_F^2\\
&=& \left\|\vec{X}_\ell-\sum_{j=1}^{M}\phi_j \nonumber \left[\vec{A}_\ell\right]_{j,:} \right\|_F^2\\
&=& \left\| \left( \vec{X}_\ell-\sum_{i \not= j}\phi_i \left[\vec{A}_\ell\right]_{i,:} \nonumber \right)-\phi_j \left[\vec{A}_\ell\right]_{j,:} \right\|_F^2\\
&\doteq& \left\| \vec{E}_\ell^j-\phi_j \left[\vec{A}_\ell\right]_{j,:} \right\|_F^2,
\end{eqnarray}
\noindent where $\vec{E}_\ell^j$ denotes the total representation error for all class $\ell$ signals when $\phi_j$ is removed. Hence, a large value of $\vec{E}_\ell^j$ indicates that the contribution of $\phi_j$ to the representation of class $\ell$ signals is large. We then define a ``representation error measure'' $m_{re}: \; \{1,2,\cdots,M\} \rightarrow \mathbb{R}_0^+$ by
\begin{equation}
m_{re}(j) \doteq \frac{r_{\ell_j^+}^j-r_{\ell_j^*}^j}{r_{\ell_j^+}^j},
\end{equation}
\noindent where $r_\ell^j \doteq \vec{E}_\ell^j / n_\ell$, for $\ell=1,2,\cdots,k$, $j=1,2,\cdots,M$.

Here again $0 \leq m_{re}(\cdot) \leq 1$, and an atom $\phi_j$ is said to be discriminative (for class $\ell_j^+$) with respect to this measure if and only if $m_{re}(j)>0$. In such a case, the value of $m_{re}(j)$ quantifies the corresponding degree of discriminability.
%--
\subsubsection{A combined discriminative measure}
Each one of the three previously defined measures takes into account different properties related to the discriminability of each one of the atoms (in a given dictionary). It is then reasonable to think of a measure that appropriately combines all three of them. With that in mind, given two positive parameters $\alpha$ and $\beta$, with $\alpha+\beta \leq 1$, we define the function $m_{\alpha,\beta}: \; \{1,2,\cdots,M\} \rightarrow \mathbb{R}_0^+$ as
\begin{eqnarray}\label{function_m}
m_{\alpha,\beta}(j) \doteq \alpha \, m_{af}(j)+ \beta \, m_{cm}(j)+ (1-\alpha-\beta) \, m_{re}(j).
\end{eqnarray}

We shall refer to $m_{\alpha,\beta}$ as the ``combined discriminative measure''. Clearly, as $\alpha$ and $\beta$ vary between $0$ and $1$, (\ref{function_m}) exhausts all possible convex combinations of the three single measures $m_{af}$, $m_{cm}$ and $m_{re}$. A challenging problem, on which we shall shed some light in Section \ref{hyp_param_tuning}, consists precisely of finding the ``optimal'' pair of parameters $(\alpha^*,\beta^*)$ leading to the best recognition rate, for a given problem.
%---
\subsection{Dictionary learning algorithm}\label{ddl}
Supervised dictionary learning methods have observed great interest in recent years. Implementations of these methods were originally focused on efficiently learning simple dictionaries (unstructured) that incorporate information of ``discriminability'' (in terms of signal classification) in their optimization process. This information can be introduced to the learning model by considering different discriminative criteria \cite{lee_efficient_2007,mairal_online_2010,wang_locality-constrained_2010}. The most commonly used criteria are the so called ``softmax'' cost function \cite{zhang_learning_2009}, Fisher criterion \cite{huang_sparse_2006} and linear predictive classification error \cite{zhang_discriminative_2010,pham_joint_2008}, to name just a few.

Although there exist several ways to simultaneously optimize both a dictionary, i.e. to solve a representation learning problem, and a classifier, i.e. to find a solution to a classification problem, a very often used strategy consists simply of dividing that problem into two sub-problems \cite{mairal_discriminative_2008,zhang_learning_2009}. Hence, it is possible to use all existing traditional dictionary learning techniques, such as MOD and KSVD, and therefore train a single classifier at a later stage. Our proposal is based precisely on this strategy but introducing class information in the dictionary learning stage. For that, we propose a new method for multi-class structured dictionary learning called ``Discriminant Atom Selection KSVD'' (DAS-KSVD) in which we use the proposed discriminative measure $m_{\alpha,\beta}$ to efficiently select class-specific discriminant atoms from some given ``auxiliary'' dictionaries to iteratively construct a structured one. The DAS-KSVD method aims at building a structures dictionary $\Phi_D^{(I)}$ by stacking side-by-side $k$ sub-dictionaries $\Phi_\ell$, each one of size $N \times I$, for all $\ell=1,2,\cdots,k$, $\Phi_D^{(I)} \doteq [\Phi_1 \; \Phi_2 \; \cdots \; \Phi_k]$. It is $N \times n$ signal matrix $\vec{X}_{trn} \doteq \left[ \vec{X}_1 \; \vec{X}_2 \; \cdots \; \vec{X}_k \right]$,important to point out that each sub-dictionary $\Phi_\ell$ contains atoms that are discriminative, in terms of $m_{\alpha,\beta}$, for class $\ell$ signals.

We now proceed to describe the building steps of the proposed DAS-KSVD method in more detail (Algorithm \ref{dksvd_algorithm}). Here, and in the sequel, we shall consider the vectors $\vec{x}_1,\vec{x}_2,\cdots,\vec{x}_n$ as realizations of a particular $N$-dimensional random vector $\mathcal{X}$.  Given an $N \times n$ signal matrix $\vec{X}_{trn} \doteq \left[ \vec{X}_1 \; \vec{X}_2 \; \cdots \; \vec{X}_k \right]$, composed of $n=\sum_{\ell=1}^{k}n_\ell$ samples, the required sparsity level $q$, the redundancy factor $r_f$, the number $t$ of class $\ell$ training signals, $t \ll n_\ell$, the number of iterations $I$ and the class label vector $\vec{c}$, the proposed algorithm begins by assigning an initial uniform probability distribution $p_0$ over $\vec{X}_{trn}$ so $p_0(i)=1/n$, for all $i$ (Alg. 1, line 2). The value of $p_0(i)$ is the probability that a training signal $\vec{x}_i$ is selected from $\vec{X}_{trn}$ in order to construct a new sampled ``learning'' matrix $\vec{X}_{lrn}$ that is used specifically for learning the initial dictionary $\Phi$. Additionally, if a certain training signal $\vec{x}_i$ is used for learning the dictionary $\Phi$ in a particular iteration, then it is desirable that such a signal be less likely than the other ones in the following iterations. Hence, promoting diversity in this way one might think that the final learned atoms are capable of highlighting different intrinsic properties of the training data.

\begin{figure}[h!]
	\centering
	\begin{overpic}[width=13cm]{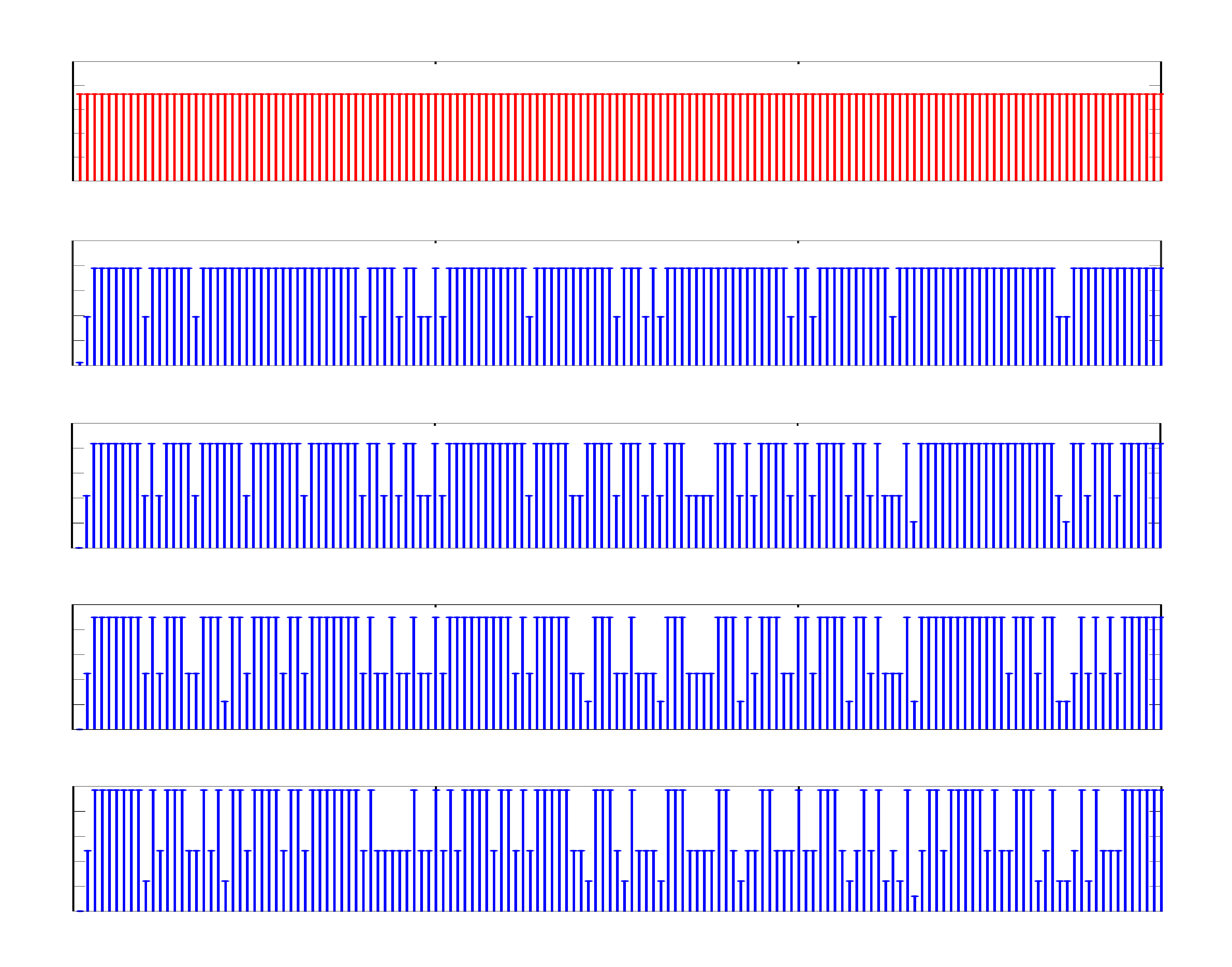}
		%\begin{overpic}[width=13cm,grid,tics=5]{images/prob_dist_various_iter.pdf}
		\put(1,37.2){\rotatebox{90}{$p_{10}$}}
		\put(1,52.6){\rotatebox{90}{$p_5$}}
		\put(1,67.3){\rotatebox{90}{$p_1$}}
		\put(1,22.5){\rotatebox{90}{$p_{15}$}}
		\put(1,7.6){\rotatebox{90}{$p_{20}$}}
		\put(6,74.2){\small $\times 10^{-5}$}
		\put(6,59.5){\small $\times 10^{-5}$}
		\put(6,44.8){\small $\times 10^{-5}$}
		\put(6,30){\small $\times 10^{-5}$}
		\put(6,15.5){\small $\times 10^{-5}$}
		\put(3.9,3.7){\small 0}
		\put(3.9,7.7){\small 1}
		\put(3.9,11.7){\small 2}
		\put(3.9,18.5){\small 0}
		\put(3.9,22.5){\small 1}
		\put(3.9,26.5){\small 2}
		\put(3.9,33.2){\small 0}
		\put(3.9,37.2){\small 1}
		\put(3.9,41.2){\small 2}
		\put(3.9,48){\small 0}
		\put(3.9,52){\small 1}
		\put(3.9,56){\small 2}
		\put(3.9,63){\small 0}
		\put(3.9,67){\small 1}
		\put(3.9,71){\small 2}
		\put(34,2){\small 50}
		\put(63,2){\small 100}
		\put(92.5,2){\small 150}
		\put(46,0){Samples ($\vec{x}_i$)}
	\end{overpic}
	\caption{Data probability distributions for five different iterations of the proposed algorithm.}
	\label{fig: prob dist diff iter}
\end{figure}
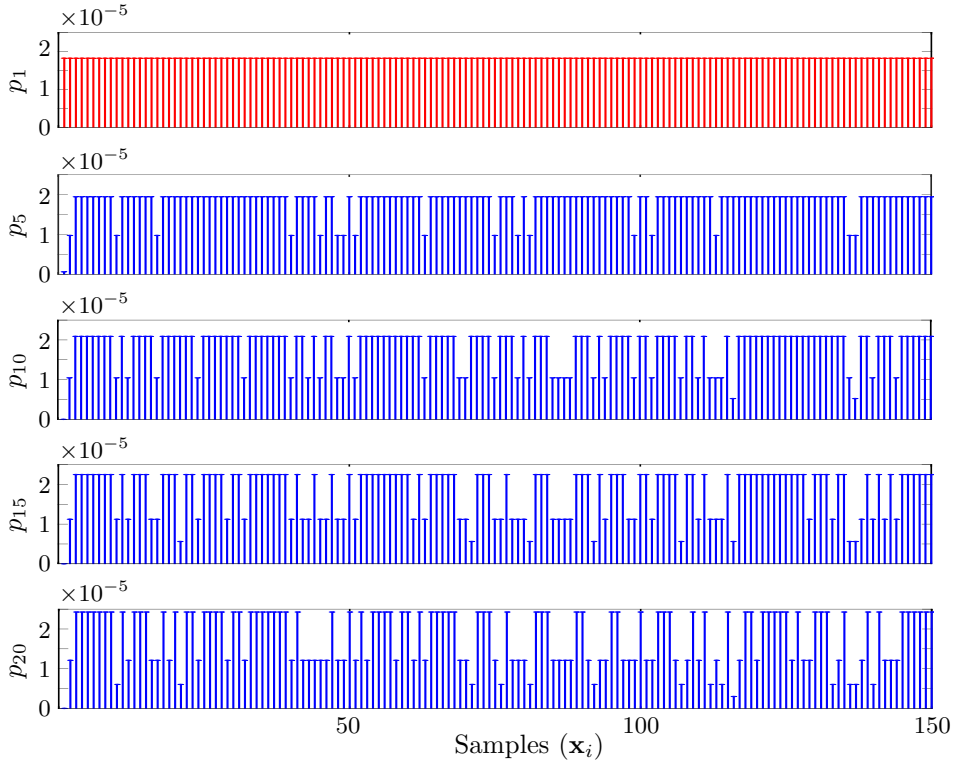

The iterative process of this algorithm (Alg. 1, lines from 3 to 10) begins by statistically sampling $t$ samples (note that $t \ll n_\ell$, for instance 10 times smaller) from each class $\ell$ signal matrix $\vec{X}_\ell$. As a result of such a sampling process, a matrix $\vec{X}_{lrn}$ of size $N \times (t*k)$ is built (Alg. 1, line 4). Also, to compute the distribution $p_{l+1}$ from both $p_l$ and $\vec{X}_{lrn}$, we multiply the value of $p_l(i)$ by a non-negative number $\tau_1<1$ if (and only if) $\vec{x}_i$ has been selected, i.e. $p_{l+1}(i)=p_l(i)\tau_1$ (in that case $p_{l+1}(i)<p_l(i)$). Otherwise $p_l(i)$ is left unchanged. It is important to point out that an appropriate normalization of these weights forcing them to sum one is needed. \figurename{ \ref{fig: prob dist diff iter}} shows graphic representations of five probability distributions $p_l$, for $l=\{1,5,10,15,20\}$. It can be observed that, at the first iteration, all samples have the same probability to be selected. In addition, see that the probability value of most samples decreases as the iteration order increases.

In order to increase robustness, all training signals used to learn the dictionary $\Phi$ (Alg. 1, line 5) are also degraded by incorporating an additive zero-mean Gaussian noise $\epsilon_{l,i}$ whose magnitude increases proportionally according to the iteration level. The magnitude of the noise is updated by $\epsilon_{l,i}=l\sigma_i\tau_2$, where $\sigma_i$ is the variance of $\vec{x}_i$ and $\tau_2$ is a (prescribed) non-negative number, $\tau_2<1$. For instance, the magnitude of the noise associated to the signal $\vec{x}_1$ at iteration 5 will be $\epsilon_{5,1}=5\sigma_1\tau_2$. It is important to point out however that, the first iteration ($l=1$) of the proposed learning algorithm leaves the original image undegraded. On the other hand, the dictionary $\Phi$ is learned by means of the traditional unsupervised KSVD algorithm \cite{aharon_ksvd_2006}. Then the sparse matrix $\vec{A}_{lrn}$ is obtained by applying the previously mentioned OMP algorithm (Alg. 1, line 6). The reason for having chosen this pursuit algorithm is because it guarantees convergence to the projection of each one of the signals into the span of the dictionary atoms, in no more than $q$ iterations leaving the rest of the coefficients equal to zero.

%---

\begin{figure}[h!]
	\centering
	\begin{overpic}[width=5in]{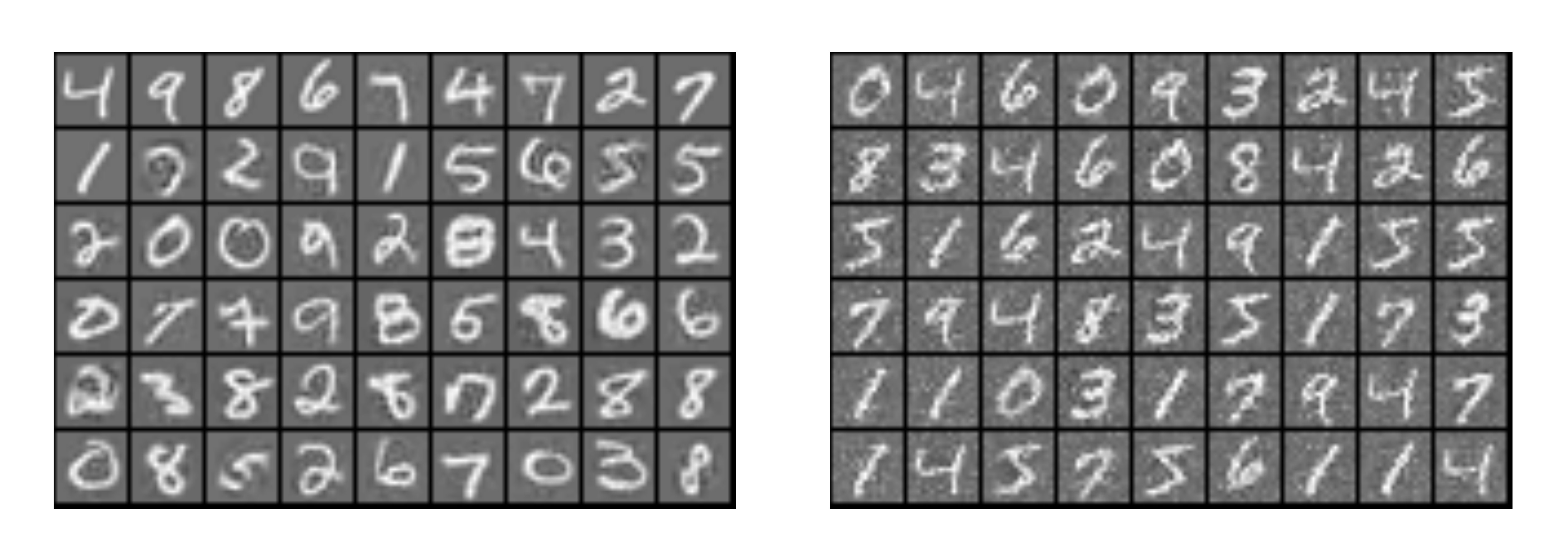}
		%\begin{overpic}[width=5in,grid,tics=5]{images/dict_learned_ksvd_iter_1_and_20.pdf}
		\put(14.5,0){Iteration 1 ($l=1$)}
		\put(62.6,0){Iteration 20 ($l=20$)}
	\end{overpic}
	\caption{Some atoms of the dictionary $\Phi$ for two different iterations of the DAS-KSVD algorithm. Iteration 1 (left) and iteration 20 (right).}
	\label{fig: dict atoms iters 1 and 20}
\end{figure}

As previously mentioned, at the beginning of each iteration, the standard unsupervised KSVD algorithm was used to learn a dictionary $\Phi$ of size $256 \times 256$. Note that this dictionary learning stage does not take into account any information concerning class membership. Additionally, the sampled subset of $t*k$ signals used to learn the dictionary was appropriately degraded by incorporating additive Gaussian noise with different magnitudes. Left and right sides of \figurename{ \ref{fig: dict atoms iters 1 and 20}} show examples of atoms coming from the dictionary $\Phi$ that were learned at iterations 1 and 20, respectively. It can be seen that, at the first iteration, the dictionary is learned by means of noise-free input signals. On the other hand, the dictionary learned at iteration 20 still preserves the structure of the handwritten digits on a blurred background.
%---

% add figure here (some atoms taken from the dictionary)!

The proposed discriminative approach consists of optimizing and using the new combined discriminative measure $m_{\alpha,\beta}$ for selecting the most discriminative atoms of $\Phi$ for each one of the $k$ classes (Alg. 1, lines from 7 to 9). As explained in Section \ref{multi_class_disc_inf}, the value of $m_{\alpha,\beta}(j)$ corresponds to the degree of discriminability of the atom $\phi_j$ for one (and only one) class, which is denoted by $\ell_j^+$. Note that the process of selecting the most discriminative atoms carries a serious trouble since the problem of finding the optimal pair of parameters $(\alpha^*,\beta^*)$ is very challenging. For more details about the tuning of that pair of parameters, we refer the reader to Section \ref{hyp_param_tuning} and Appendix A. Also, the construction of the sub-dictionary $\Phi_d$ (Alg 1, line 8) basically consists of taking one-by-one the most discriminative atoms of $\Phi$ for each one of the $k$ classes and stacking them side-by-side. In the case that there is more than one $\ell_j^+$ class-related candidate complying with the proposed discriminative criterion, $\phi_j$ is defined as the atom that maximizes all possible values of $m_{\alpha^*,\beta^*}$. Otherwise, in case that $\Phi$ lacks of discriminative atoms, the signal selection process (Alg. 1, line 4) is restarted.

\begin{algorithm}[h!]
	\caption{Pseudocode of the new DAS-KSVD method}\label{dksvd_algorithm}
	\begin{algorithmic}[1]
		\Procedure{das-ksvd}{$\vec{X}_{trn},q,r_f,t,I,\vec{c}$}
		\State $p_0(i)=1/n$, for all $i$
		\For{$l \gets 0, I-1$} % number of iterations
		\State $\left[ \vec{X}_{lrn},p_{l+1} \right] \gets $~\Call{SampleData}{$\vec{X}_{trn},t,p_l,l$}
		\State $\Phi \gets $~\Call{Ksvd}{$\vec{X}_{lrn},r_f,q$}
		\State $\vec{A}_{lrn} \gets $~\Call{OMP}{$\vec{X}_{lrn},\Phi,q$}
		\State $m_{\alpha^*,\beta^*} \gets $~\Call{DiscMeasure}{$\vec{A}_{lrn},\vec{c},q$}
		\State $\Phi_d \gets $~\Call{GetAtoms}{$\Phi,m_{\alpha^*,\beta^*}$}
		\State $\Phi_D^{(i)} \gets $~\Call{SaveAtoms}{$\Phi_d$}
		\EndFor
		\State \textbf{return} $\Phi_D^{(I)}$
		\EndProcedure
	\end{algorithmic}
\end{algorithm}
%\subsection{Classifiers}\label{classifiers}
%This section presents a description of the classifiers that were used to asses the proposed method.
%---
\subsection{Classifier} \label{mlp}
In this work a Multilayer Perceptron (MLP) neural network is used in order to assess the proposed method. The MLP neural network is one of the most popular classes of neural networks whose architecture consists of a fully connected assembly of single artificial neurons. The MLP neural network is typically comprised by an input layer, one (or more) hidden layers and an output layer \cite{haykin_neural_1998}. The inputs (features) are processed layer-by-layer moving forward through the network. Each artificial neuron receives one (or more) inputs from its preceding nodes, processes the information and produces an output that is transmitted to the next node. The output of each neuron is reached by applying an activation (transfer) function (linear or not) to the weighted sum of the inputs plus a bias term. More precisely, the output of a neuron $y_j$ is defined as
\begin{equation}
y_j=f\left(\sum_{i=1}^d\omega_{ji}x_i+\omega_{j0}\right)=f\left(\sum_{i=0}^d\omega_{ji}x_i\right),
\label{eq_net}
\end{equation}
\noindent where the transfer function is denoted by $f(\cdot)$, and the weights that connect the $i^{\textrm{th}}$-input to the $j^{\textrm{th}}$-neuron for a given layer is represented by $\omega_{ji}$.

Since the MLP neural network training process is supervised, the desired outputs (labels) are required. The most popularly used method for training MLP neural networks is the back-propagation algorithm \cite{rumelhart_learning_1986}. This algorithm iteratively adjusts the synaptic weights in the network by minimizing a given measure which quantifies the difference between the current output vector and the desired one.

\section{Experiments}\label{sec: experiments}
In this section we present a brief description regarding the experimental setup. Additionally, we make a brief recall of the evaluation metric used for assessing the proposed dictionary learning method. Finally, we comment on appropriate ways for tuning the parameters.
%---

\subsection{Experimental setup}
As mentioned above, the performance of the new DAS-KSVD method is evaluated using standard partitions for training and testing of the MNIST database. Although it is not a requirement, our experiments were performed by using a balanced set of training and validation samples. For that, subsets consisting of 4,000 and 1,000 images for each one of the classes coming from the standard partition of the training dataset were randomly chosen. Hence, new training and validation matrices ($\vec{X}_{trn}$ and $\vec{X}_{val}$) comprised by 40,000 and 10,000 samples, respectively, were built. It is important to point out however that, the standard partition of the testing dataset $\vec{X}_{tst}$ of size $256 \times 10,000$ was left unchanged.

It becomes appropriate to mention that the matrix $\vec{X}_{trn}$ was used both for dictionary learning and training the MLP neural network while the matrix $\vec{X}_{val}$ was used for testing the MLP neural network as well as for parameters tuning. Furthermore, the matrix $\vec{X}_{tst}$ was only taken into account for performing the final test.

We shall now proceed to describe the parameter settings for the DAS-KSVD method that were used in the experiments. We evaluated the effect that produces the size of $\Phi_D^{(I)}$ in the final recognition rate. For that, we have considered four structured dictionaries denoted by $\Phi_D^{(5)}$, $\Phi_D^{(10)}$, $\Phi_D^{(15)}$ and $\Phi_D^{(20)}$ which are composed by 50, 100, 150 and 200 atoms, respectively. Hence, the DAS-KSVD algorithm was run 20 iterations, i.e. $I=20$.

\subsection{Evaluation metric}
Overall accuracy rate constitutes one of the most popular performance measures used to assess pattern recognition-related methods. The accuracy measure (Acc) is defined as the proportion of correctly predicted testing samples. Let $n$ the number of testing samples, $\lambda_i$ and $\hat{\lambda}_i$ the label and prediction, respectively, regarding $\vec{x}_i$ and $\delta(x,y)$ the well known delta function whose output is true (one) if $x=y$ and false (zero) otherwise. The Acc measure is defined as:
\begin{equation}
\textrm{Acc}=\frac{1}{n}\sum_{i=1}^{n}\delta(\lambda_i,\hat{\lambda}_i).
\end{equation}
%---
\subsection{Parameters tuning}\label{hyp_param_tuning}
Although the pursuit for discriminative atoms is perhaps one of the most challenging issues to be addressed in this work, finding optimal pair of parameters ($\alpha^*,\beta^*$) leading to the best recognition rate is also a very difficult task. However, the problem of finding that optimal pair of parameters strongly depends on the application under study. For that reason, we propose applying the well known and widely used ``grid search'' method for parameter optimization. For more details regarding grid search method, we refer the reader to Appendix A. In what follows, the final choice of the remaining parameters of the proposed algorithm are described.

At each iteration of the proposed DAS-KSVD method, one (and only one) discriminant atom for each one of the $k$ classes is selected. Hence, each iteration of this method generates $k$ discriminative atoms and therefore, if the algorithm is configured to perform $I$ iterations, then the final structured dictionary will be composed by $I*k$ discriminant atoms. In order to explore the effect of the final structured dictionary size, the experiments were performed by considering a total of 20 iterations, i.e. $I=20$. Thus, the final discriminative dictionary $\Phi_D^{(I)}$ is composed by 200 atoms (assuming $k=10$). On the other hand, the number of samples for each class used to learn the full dictionary was set to $t=500$.

As described in Section \ref{ddl}, $\tau_1$ and $\tau_2$ are two parameters ($0 \leq \tau_1,\tau_2 < 1$) that need to be adjusted and fixed. Several trials were performed in order to obtain appropriate values for those parameters. A value of $\tau_1=0.5$ was finally selected and used in our experiments. Additionally, it was found that a value of $\tau_2=0.1$ presented the best trade-off between image degradation and iteration order. 

The standard KSVD algorithm starts by performing a random selection of 256 samples coming from the learning signal matrix $\vec{X}_{lrn}$. Note that the redundancy factor ($r_f$) used for constructing the dictionary is equal to one, i.e. $M=N=256$. Also, the maximum number of KSVD iterations was fixed to 50 in the code. It is also well known that the KSVD algorithm internally computes sparse codes representing each one of all involved signals. These codes were obtained by means of OMP algorithm. To establish an appropriate sparsity level, a great variety of sparse solutions were tested. It was found that a sparsity degree of 20\% presents the best trade-off between discriminability and representativity of all signals.

The MLP neural network training process was performed using back-propagation method. This algorithm was optimized minimizing the Mean Squared Error (MSE) function through Scaled Conjugate Gradient (SCG) method. Also, the output of each neuron was determined by applying a saturating linear transfer function. Additionally, the structure of the MLP neural network was configured such that the sizes of its hidden and input layers are equal.
%---
\section{Results and discussion}\label{sec: results and discussion}
As already explained above, the matrices denoted by $\vec{A}_{trn}$ and $\vec{A}_{val}$ provide the sparse representations of $\vec{X}_{trn}$ and $\vec{X}_{val}$, respectively, in terms of a dictionary $\Phi$ through $\vec{X}_{trn}=\Phi \vec{A}_{trn}$ and $\vec{X}_{val}=\Phi \vec{A}_{val}$. Also, the feature vectors $\vec{a}_i$ comprising the matrices $\vec{A}_{trn}$ and $\vec{A}_{val}$ were used as inputs for training and testing, respectively, the MLP neural network. The final test was performed by taken into account the standard partition of the testing dataset $\vec{X}_{tst}$ and each one of the previously learned structured dictionaries $\Phi_D^{(I)}$. The matrix $\vec{A}_{tst}$ was obtained by means of the OMP algorithm. Also, the inputs of the already trained MLP neural networks are the feature vectors $\vec{a}_i$ coming from $\vec{A}_{tst}$ and, moreover, the outputs of these networks are evaluated to compute the final accuracy. In addition, structured dictionaries composed by 50, 100, 150 and 200 discriminant atoms were evaluated. Mean and standard deviation of the classification results over 10 rounds were found to be 94.87\% ($\pm$ 0.33\%), 94.79\% ($\pm$ 0.30\%), 94.36\% ($\pm$ 0.27\%) and 91.25\% ($\pm$ 0.63\%) for feature vector sizes of 200, 150, 100 and 50, respectively. Also, Table 1 presents a comparative summary of the best recognition rates yielded by MLP neural networks trained using as input the matrix $\vec{A}_{trn}$ obtained by taken into account each one of the evaluated structured dictionaries. Also, details regarding the required number of weights of the MLP neural network for each one of such dictionaries are included. It is important to point out that these results were obtained by considering a fixed hidden layer size coinciding with the input feature vector size. Maximal accuracy rates of 96.2, 95.9, 95.0 and 92.2 were obtained for feature vector sizes of 200, 150, 100 and 50, respectively. Hence, results show that ``discriminative'' feature vectors of length 200 are the best option for handwritten digits recognition. On the other hand, the last column of Table 1 shows the total number of weights required to train each one of the MLP neural networks.
\begin{table}[h!]
	\begin{center} \small
		\caption{Best recognition rates on the test set yielded by MLP neural networks using DAS-KSVD feature vectors as well as the number of weights required for their training.}
		\renewcommand{\arraystretch}{1.5}
		\begin{tabular}{*{4}{c}}
			\noalign{\smallskip}\hline\noalign{\smallskip}
			Dictionary & Classifier & Acc (\%) & Number of weights \\
			\noalign{\smallskip}\hline\noalign{\smallskip}
			$\Phi_D^{(5)}$ & MLP-50-50-10 & 92.23 & 3,060  \\
			$\Phi_D^{(10)}$ & MLP-100-100-10 & 95.03 & 11,110 \\
			$\Phi_D^{(15)}$ & MLP-150-150-10 & 95.90 & 24,160 \\
			$\Phi_D^{(20)}$ & MLP-200-200-10 & \textbf{96.20} & 42,210 \\
			\noalign{\smallskip}\hline\noalign{\smallskip}
		\end{tabular}
		\label{tab: compar recogn rates four dict sizes}
	\end{center}
\end{table}

Lecun \textit{et al.} \cite{lecun_gradient-based_1998} tested several configurations of one-hidden layer fully connected MLP neural networks trained for handwritten digit recognition. One of them consists of directly using the original (raw) data, i.e. without tacking into account any signal pre-processing or feature selection, as input of the classifier. Thus, vectors containing 784 features corresponding to images of size $28 \times 28$ were used as inputs of the classifier. The first two rows of \tablename{ 2} shows maximal percentages of accuracy rates (Acc) yielded by MLP neural networks with 300 (MLP-784-300-10) and 1000 (MLP-784-1000-10) neurons in their hidden layer. The number of training weights for each one of the networks are also included in the last column. Accuracy rates on the standard test partition of 95.3\% and 95.5\% were yielded by MLP neural networks with 300 and 1000 hidden neurons, respectively. It can be observed that, as a result of increasing the number of hidden neurons (from 300 to 1000), a slight improvement in the result was achieved. Also, the number of weights of the network has increased from 238,510 to 795,010, which represent an increment of 333\%.

\begin{table}[h!] 
	\begin{center} \small 
		\caption{Best recognition rates on the test set yielded by MLP neural networks using DAS-KSVD feature vectors as well as the ones derived from dictionaries learned with the other three evaluated methods.}
		\begin{tabular}{*{4}{c}}
			\noalign{\smallskip}\hline\noalign{\smallskip}
			Method & Classifier & Acc (\%) & Number of weights \\
			\noalign{\smallskip}\hline\noalign{\smallskip}
			\multirow{2}{*}{Raw data \cite{lecun_gradient-based_1998}} & MLP-784-300-10 & \textbf{95.3} & \textbf{238,510} \\
			& MLP-784-1000-10 & 95.5 & 795,010 \\
			%& (dist) MLP-784-300-10 & \textbf{96.4} & 238,510 \\
			%& (dist) MLP-784-1000-10 & 96.2 & 795,010 \\
			\noalign{\smallskip}\hline\noalign{\smallskip}
			\multirow{5}{*}{DAS-KSVD} & MLP-200-50-10 & \textbf{95.3} & \textbf{10,560} \\
			& MLP-200-100-10 & 96.1 & 21,110 \\
			& MLP-200-200-10 & 96.2 & 42,210 \\
			& MLP-200-300-10 & 96.4 & 63,310 \\
			& MLP-200-1000-10 & \textbf{96.7} & \textbf{211,010} \\
			\noalign{\smallskip}\hline\noalign{\smallskip}
			\multirow{5}{*}{KSVD \cite{aharon_ksvd_2006}} & MLP-200-50-10 & 93.5 & 10,560 \\
			& MLP-200-100-10 & 92.8 & 21,110 \\
			& MLP-200-200-10 & 92.3 & 42,210 \\
			& MLP-200-300-10 & 92.8 & 63,310 \\
			& MLP-200-1000-10 & 92.7 & 211,010 \\
			\noalign{\smallskip}\hline\noalign{\smallskip}
			\multirow{5}{*}{LC-KSVD2 \cite{jiang_label_2013}} & MLP-200-50-10 & 91.8 & 10,560 \\
			& MLP-200-100-10 & 91.9 & 21,110 \\
			& MLP-200-200-10 & 92.0 & 42,210 \\
			& MLP-200-300-10 & 92.1 & 63,310 \\
			& MLP-200-1000-10 & 92.3 & 211,010 \\
			\noalign{\smallskip}\hline
		\end{tabular}
		\label{tab_compar_feature_sel}
	\end{center}
\end{table}

Table 2 also shows a comparative summary of the results yielded by MLP neural networks with a reduction in the dimension of the feature vectors. For that, the proposed DAS-KSVD method was used for obtaining feature vectors of length 200. As shown in \tablename{ \ref{tab: compar recogn rates four dict sizes}}, structured dictionaries composed by 200 discriminative atoms ($\Phi_D^{(20)}$) are the best option for handwritten digit recognition. Clearly, the use of small dimensional feature vectors produce a significant dimension reduction \textcolor{black}{but retaining discriminative information} and therefore, the computing time required for classification is reduced. Thus, the number of input units of the MLP neural network was reduced (from 784 to 200) in 74.49\% compared with those required by the original raw data. The table shows the average over 10 rounds of accuracy rates yielded by MLP neural networks with 200 input units while varying the number of hidden neurons from 50 to 1000. The last column of this table shows the required number of training weights. Accuracy rates on the standard testing dataset of 96.4\% and 96.7\% were achieved by MLP neural networks with 300 (MLP-200-300-10) and 1000 (MLP-200-1000-10) hidden neurons, respectively. Additionally, the performance of MLP neural networks with 50, 100 and 200 hidden neurons were tested without showing significant improvements in the results. 

It is also important to point out that the classifier MLP-200-50-10 (DAS-KSVD method) has achieved the same recognition rate (95.3\%) as MLP-784-300-10 (Raw data) 
using a MLP neural network composed by only a 4.42\% of the required weights. It was also found that taking into account the best option that uses the original raw data as inputs of the classifier (MLP-784-1000-10), it has 795,010 training weights while DAS-KSVD method (MLP-200-1000-10) has not only 211,010 weights, but also increases a 1.2\% in the performance of the classifier. As a result of that analysis, one might think that the proposed DAS-KSVD method produces a significant dimension reduction while enhancing the overall recognition rate. Summing up, it was demonstrated that using the proposed DAS-KSVD method for dimension reduction undoubtedly enhances the recognition rate of MLP neural networks.

We have compared the performance of the new DAS-KSVD method with the standard KSVD method as well as with the discriminative-based LC-KSVD2 method. It can be observed from Table 2 that the proposed DAS-KSVD method outperform all the others showing robustness and effectiveness \textcolor{black}{with the same size of the dictionary} in the recognition of handwritten digits images coming from MNIST database. The maximum recognition rate yielded by the DAS-KSVD method was 96.7\% which clearly outperforms those yielded by both KSVD (93.5\%) and LC-KSVD2 (92.3\%) methods.

We have also evaluated the statistical significance of the results presented in Table 2 by computing the probability that the DAS-KSVD method yields better recognition rates than all the other evaluated methods $(P(\epsilon_{ref}<\epsilon))$. In order to perform this test we assumed the statistical independence of the classification errors for each image and we approached the error's Binomial distribution by means of a Gaussian distribution. This is possible because we have a sufficiently high number of testing samples (10,000). \textcolor{black}{In this way, for 95.5\% and 96.7\% corresponding to recognition rates yielded by ``Raw data'' that produced the best performance among all methods considered in the experiments and the new proposed one (DAS-KSVD), respectively, we have that $P(\epsilon_{ref}<\epsilon)>0.9999$.}

\section{Conclusions}\label{sec: conclusions}
In this work, both a new discriminative measure and a novel method for learning structured dictionaries for multi-class classification problems were introduced. This new measure is capable of efficiently quantifying the degree of discriminability of each one of the atoms in a particular dictionary. The use of such a measure gave rise to what we called the Discriminant Atom Selection KSVD (DAS-KSVD) method for dictionary learning. The method was tested with a widely used database for handwritten digit recognition and compared with three state-of-the-art classification methods. Experimental results showed that DAS-KSVD significantly outperforms the other three methods achieving good recognition rates and additionally, reducing the computational cost of the classifier.

Clearly, there is much further room for improvements. In particular, future research lines include the evaluation of our learning method with other well known databases, more analysis of the combined discriminative measure as well as the study of its properties and the exploration of new deep structures. %Also, we plan to study the properties of the obtained learned dictionaries and also explore different ways to incorporate discriminative information into the dictionary learning process. It is also of our interest to apply these techniques to other breathing-related sleep disorder problem as well as to other problems coming from different areas such as computer vision, speech processing, etc.
%---
\section{Acknowledgments}
%All acknowledments will be placed here.
This work was supported in part by Consejo Nacional de Investigaciones Científicas y Técnicas, CONICET, through PIP 2014-2016 No. 11220130100216-CO, by Agencia Nacional de Promoción Científica y Técnica, ANPCyT, under projects PICT 2014-2627, PICT 2015-0977 and PICT 2017-4596 and by Universidad Nacional del Litoral, UNL, through projects CAI+D 500 201501 00059 LI, CAI+D 500 201501 00082 LI and CAI+D 504 201501 00036 LI ``Problemas Inversos y Aplicaciones a Procesamiento de Señales e Imágenes''.

%\newpage
%\bibliographystyle{ieeetr}
%\bibliography{bib_disc_dict_learning.bib}

\newpage
%\appendix
%\section{Grid search} \label{ap: parameters tuning}
\begin{appendices}
	\section{Grid search}
\begin{figure}[h!]
		\centering
		\begin{overpic}[width=5in]{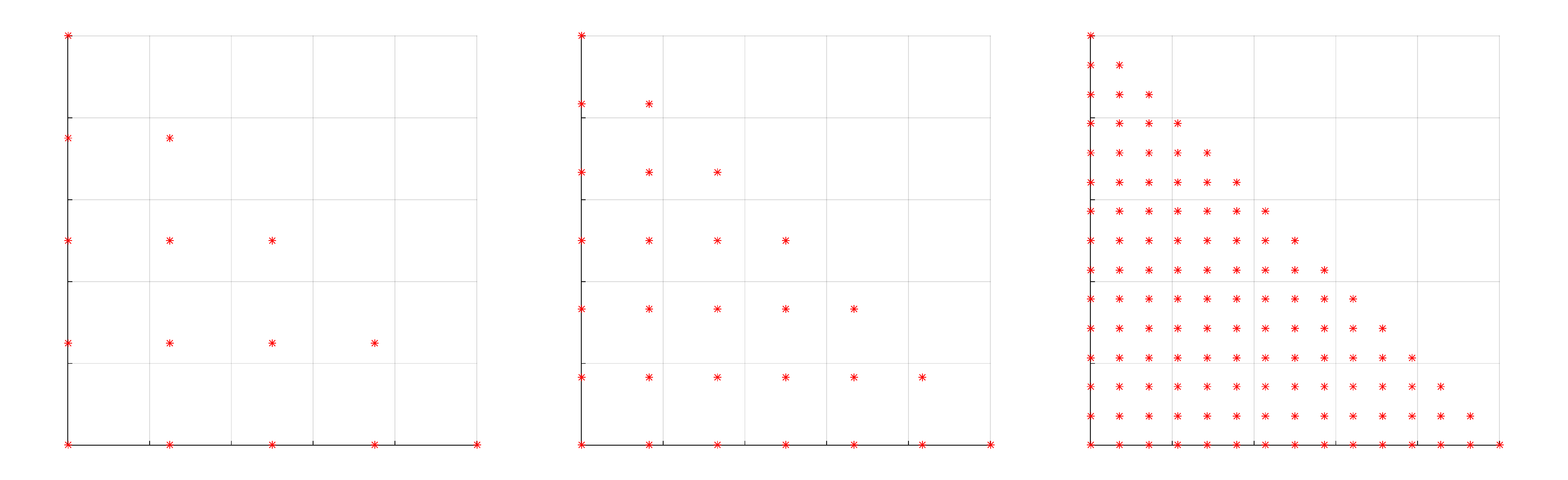}
			%\begin{overpic}[width=5in,grid,tics=5]{images/grid_search_examples.pdf}
			\put(-1,14.5){\small \rotatebox{90}{$\beta$}}
			\put(31.5,14.5){\small \rotatebox{90}{$\beta$}}
			\put(64,14.5){\small \rotatebox{90}{$\beta$}}
			\put(14,29){\small $\Delta_1 = 1/4$}
			\put(46,29){\small $\Delta_1 = 1/6$}
			\put(78,29){\small $\Delta_1 = 1/14$}
			\put(16.5,-2){\small $\alpha$}
			\put(49.4,-2){\small $\alpha$}
			\put(81.8,-2){\small $\alpha$}
			\put(3.7,-0.2){\small 0}
			\put(8,-0.2){\small 0.2}
			\put(13.2,-0.2){\small 0.4}
			\put(18.4,-0.2){\small 0.6}
			\put(23.7,-0.2){\small 0.8}
			\put(29.8,-0.2){\small 1}
			\put(36.5,-0.2){\small 0}
			\put(40.7,-0.2){\small 0.2}
			\put(46,-0.2){\small 0.4}
			\put(51.2,-0.2){\small 0.6}
			\put(56.4,-0.2){\small 0.8}
			\put(62.6,-0.2){\small 1}
			\put(69,-0.2){\small 0}
			\put(73.3,-0.2){\small 0.2}
			\put(78.4,-0.2){\small 0.4}
			\put(83.6,-0.2){\small 0.6}
			\put(88.9,-0.2){\small 0.8}
			\put(95.1,-0.2){\small 1}
			\put(1.7,5.8){\small \rotatebox{90}{0.2}}
			\put(34.5,5.8){\small \rotatebox{90}{0.2}}
			\put(67,5.8){\small \rotatebox{90}{0.2}}
			\put(1.7,11){\small \rotatebox{90}{0.4}}
			\put(34.5,11){\small \rotatebox{90}{0.4}}
			\put(67,11){\small \rotatebox{90}{0.4}}
			\put(1.7,16.2){\small \rotatebox{90}{0.6}}
			\put(34.5,16.2){\small \rotatebox{90}{0.6}}
			\put(67,16.2){\small \rotatebox{90}{0.6}}
			\put(1.7,21.4){\small \rotatebox{90}{0.8}}
			\put(34.5,21.4){\small \rotatebox{90}{0.8}}
			\put(67,21.4){\small \rotatebox{90}{0.8}}
			\put(1.7,27.6){\small \rotatebox{90}{1}}
			\put(34.5,27.6){\small \rotatebox{90}{1}}
			\put(67,27.6){\small \rotatebox{90}{1}}
		\end{overpic}
		\caption{Different possible combinations of weights.}
		\label{fig: grid search}
	\end{figure}
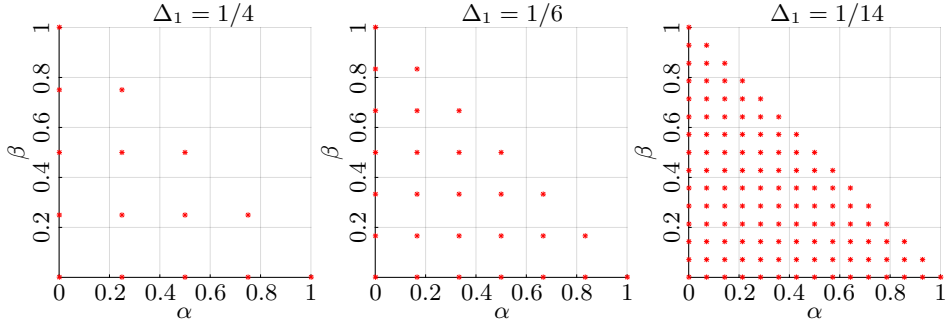

	The grid search method starts by dividing the interval $\left[0;1\right]$ into segments of length $\Delta$ and generating different combinations of the parameters $\alpha$ and $\beta$ such that $\alpha+\beta \leq 1$. This constraint suggests that the boundary of the work space coincides with a right triangle whose vertexes are the pair of parameters corresponding to $\left(0,0\right)$, $\left(1,0\right)$ and $\left(0,1\right)$. \figurename{ \ref{fig: grid search}} shows an example of the grid search method for three different values of $\Delta$. It can be observed that small values of $\Delta$ entail evaluating a large number of combinations.
	
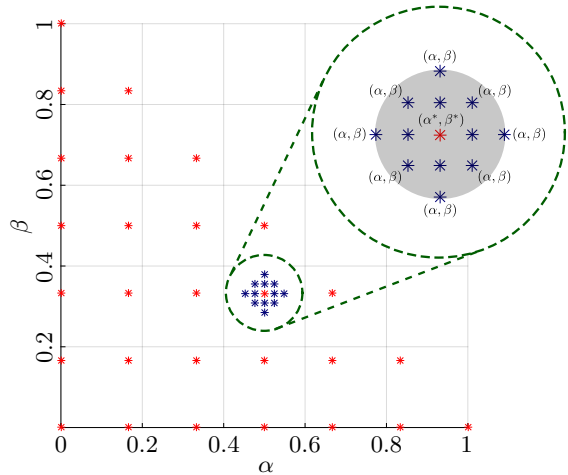
\begin{figure}[h!]
		\centering
		\begin{overpic}[width=3in]{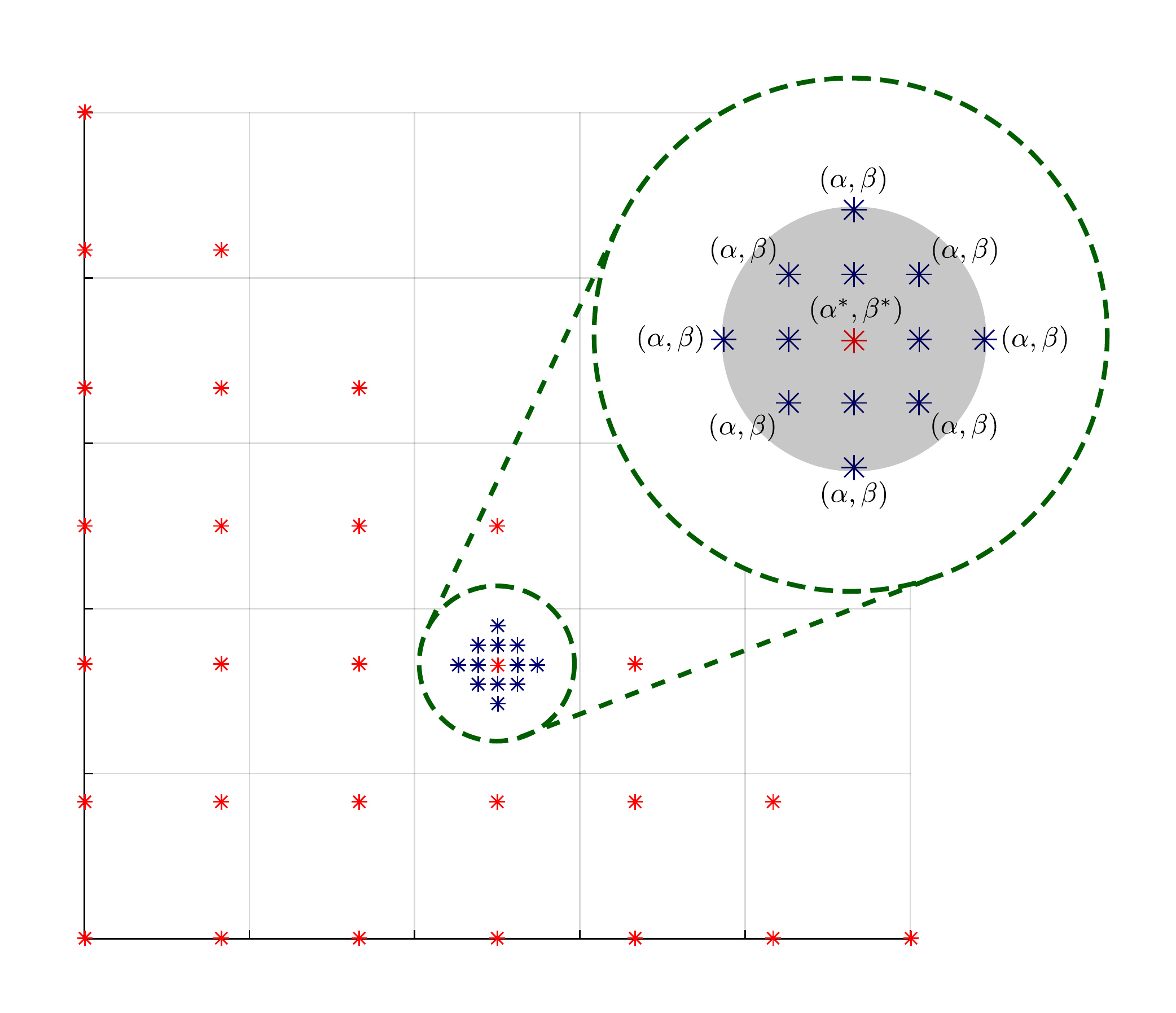}
			%\begin{overpic}[width=3in,grid,tics=5]{images/grid_search_7_zoom.pdf}
			\put(-1.5,40){\small \rotatebox{90}{$\beta$}}
			\put(41.2,-1.5){\small $\alpha$}
			\put(6.1,2){\small 0}
			\put(18.5,2){\small 0.2}
			\put(32.6,2){\small 0.4}
			\put(46.6,2){\small 0.6}
			\put(60.7,2){\small 0.8}
			\put(76.2,2){\small 1}
			\put(3,17.5){\small \rotatebox{90}{0.2}}
			\put(3,31.5){\small \rotatebox{90}{0.4}}
			\put(3,45.8){\small \rotatebox{90}{0.6}}
			\put(3,59.7){\small \rotatebox{90}{0.8}}
			\put(3,75.2){\small \rotatebox{90}{1}}
		\end{overpic}
		\caption{An example illustrating a second stage grid search.}
		\label{fig: grid search_zoom}
	\end{figure}

	In order to reduce the computational cost, we have performed a grid search of the optimal pair of parameters into two stages. The first one consists of defining and using $\Delta_1=1/6$ in order to locate potential ``regions'' in the search space where recognition rates are maximized. Also, the second stage takes into account these regions and, moreover, performs a more refined search using $\Delta_2=1/100$. In that way, each new refined region of search is established by considering all possible pair of parameters complying with $(\alpha-\alpha^*)^2+(\beta-\beta^*)^2 \leq (2\Delta_2)^2$ (see \figurename{ \ref{fig: grid search_zoom}}). This definition coincides with all $(\alpha,\beta)$ that are inside to a close disc of radius $2 \Delta_2$ centered at $(\alpha^*,\beta^*)$. 
	
	The most discriminative atoms of $\Phi$ according to the combined measure $m_{\alpha,\beta}$ were selected and taken in for building structured dictionaries. As mentioned above, the problem of finding the optimal pair of parameters $(\alpha^*,\beta^*)$ was solved by applying the grid search method. This search was initially carried out by taking into account an interval length of $\Delta=1/6$ which leads to 28 different pair of parameters. \figurename{ 7} shows a summary of the results obtained by applying the grid search method for each one of the four evaluated dictionaries. In particular, we have found that using structured dictionaries comprised by more than 5 class-related discriminative atoms, the MLP neural networks achieved good recognition rates. This figure also shows, for each one of the evaluated dictionaries, two highlighted regions denoted by $\textrm{R}_1$ and $\textrm{R}_2$ where recognition rate are maximum. Among all highlighted regions, one might think that simultaneous values of $\alpha$ and $\beta$ close to zero allow selecting the most discriminative atoms of $\Phi$. In case of using a structured dictionary comprised by 5 discriminative atoms for each one of the classes, we found that search regions $\textrm{R}_1$ and $\textrm{R}_2$ are centered at $(0.33,0.17)$ and $(0.83,0)$, respectively, and centered at $(0,0)$ and $(0.33,0.17)$, otherwise. 
	
	%---
	\begin{figure}[h!]
		\centering
		\subfigure{}
		\begin{overpic}[width=0.4\textwidth,tics=5]{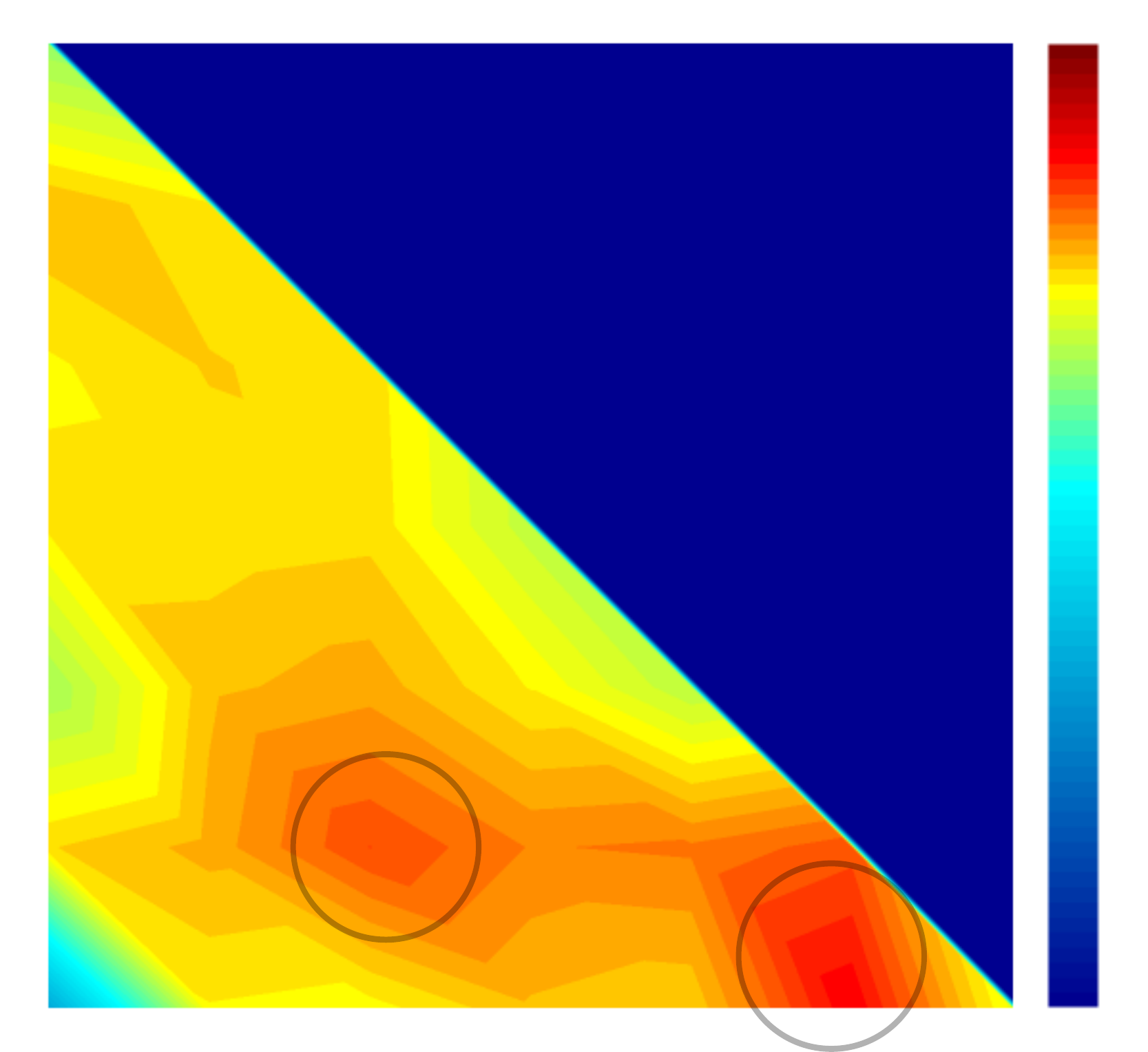}
			\put(43,90.5){$\Phi_D^{(5)}$}
			\put(3,-0.5){0}
			\put(19.8,-0.5){0.2}
			\put(36.6,-0.5){0.4}
			\put(53.2,-0.5){0.6}
			\put(69.8,-0.5){0.8}
			\put(87,-0.5){1}
			\put(-3,17){0.2}
			\put(-3,34){0.4}
			\put(-3,50){0.6}
			\put(-3,66){0.8}
			\put(1,86){1}
			\put(97,86){93}
			\put(97,2){86}
			\put(62,18){$\textrm{R}_2$}
			\put(37,27.5){$\textrm{R}_1$}
			\put(43,-4){$\alpha$}
			\put(-8,43){\rotatebox{90}{$\beta$}}
			\put(97,60){\rotatebox{-90}{Accuracy (\%)}}
		\end{overpic} \hspace{5mm}
		\subfigure{}
		\begin{overpic}[width=0.4\textwidth]{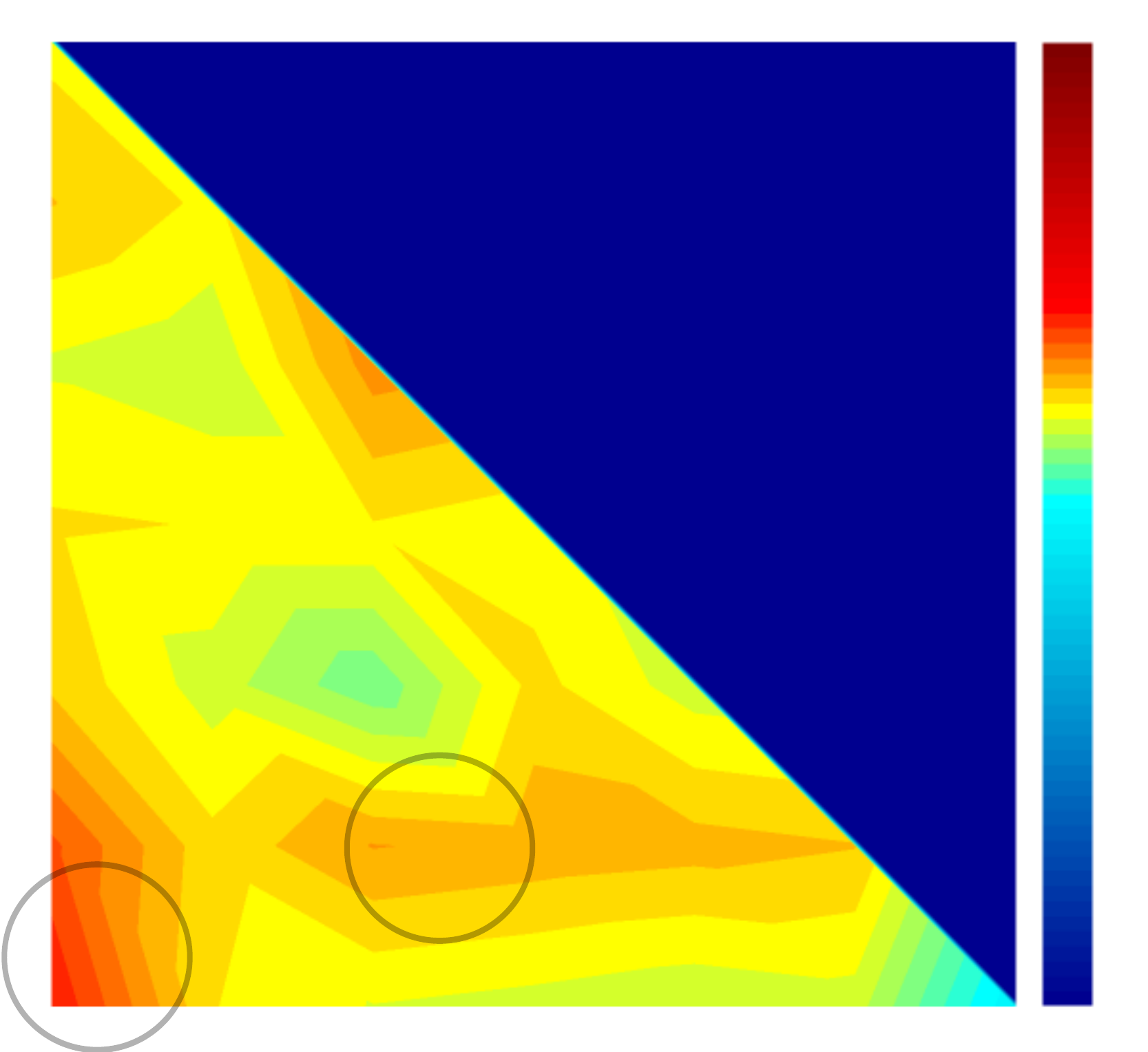}
			\put(43,90.5){$\Phi_D^{(10)}$}
			\put(3,-0.5){0}
			\put(19.8,-0.5){0.2}
			\put(36.6,-0.5){0.4}
			\put(53.2,-0.5){0.6}
			\put(69.8,-0.5){0.8}
			\put(87,-0.5){1}
			\put(-3,17){0.2}
			\put(-3,34){0.4}
			\put(-3,50){0.6}
			\put(-3,66){0.8}
			\put(1,86){1}
			\put(97,86){97}
			\put(97,2){90}
			\put(7,18){$\textrm{R}_1$}
			\put(37,27.5){$\textrm{R}_2$}
			\put(43,-4){$\alpha$}
			\put(-8,43){\rotatebox{90}{$\beta$}}
			\put(97,60){\rotatebox{-90}{Accuracy (\%)}}
		\end{overpic} \vspace{5mm}
		\subfigure{}
		\begin{overpic}[width=0.4\textwidth]{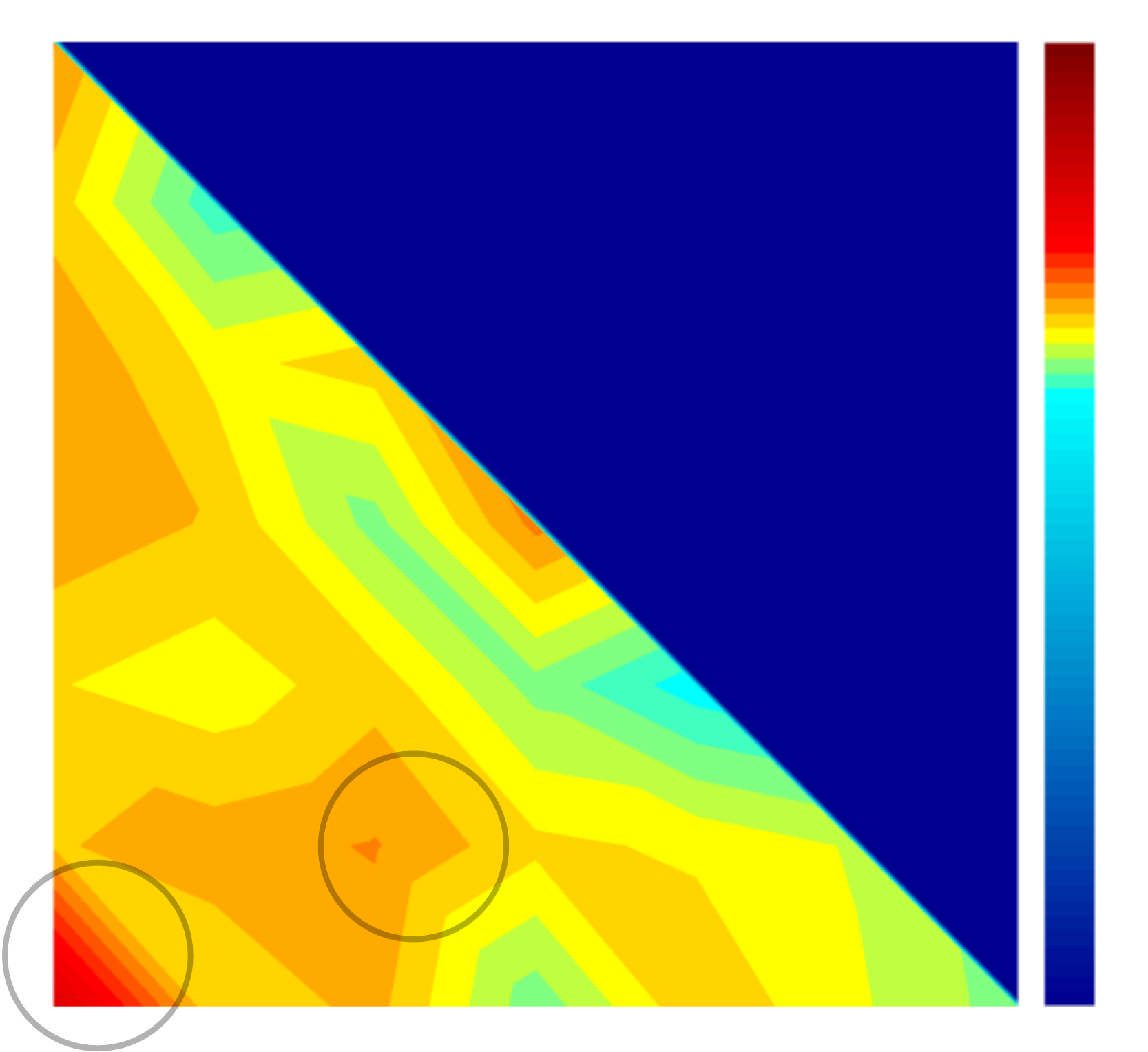}
			\put(43,90.5){$\Phi_D^{(15)}$}
			\put(3,-0.5){0}
			\put(19.8,-0.5){0.2}
			\put(36.6,-0.5){0.4}
			\put(53.2,-0.5){0.6}
			\put(69.8,-0.5){0.8}
			\put(87,-0.5){1}
			\put(-3,17){0.2}
			\put(-3,34){0.4}
			\put(-3,50){0.6}
			\put(-3,66){0.8}
			\put(1,86){1}
			\put(97,86){97}
			\put(97,2){90}
			\put(7,18){$\textrm{R}_1$}
			\put(37,27.5){$\textrm{R}_2$}
			\put(43,-4){$\alpha$}
			\put(-8,43){\rotatebox{90}{$\beta$}}
			\put(97,60){\rotatebox{-90}{Accuracy (\%)}}
		\end{overpic} \hspace{5mm}
		\subfigure{}
		\begin{overpic}[width=0.4\textwidth]{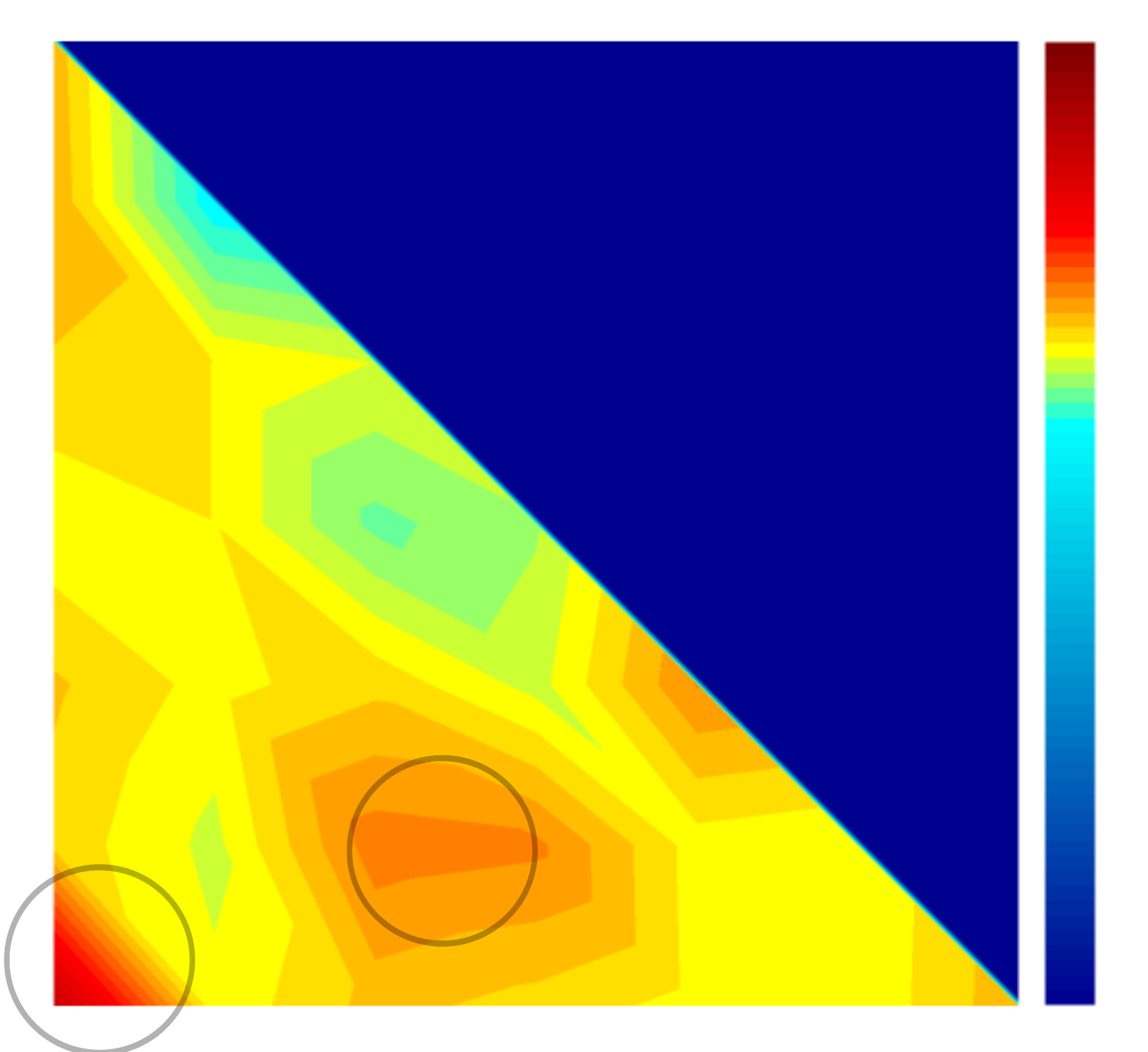}
			\put(43,90.5){$\Phi_D^{(20)}$}
			\put(3,-0.5){0}
			\put(19.8,-0.5){0.2}
			\put(36.6,-0.5){0.4}
			\put(53.2,-0.5){0.6}
			\put(69.8,-0.5){0.8}
			\put(87,-0.5){1}
			\put(-3,17){0.2}
			\put(-3,34){0.4}
			\put(-3,50){0.6}
			\put(-3,66){0.8}
			\put(1,86){1}
			\put(97,86){97}
			\put(97,2){90}
			\put(7,18){$\textrm{R}_1$}
			\put(37,27.5){$\textrm{R}_2$}
			\put(43,-4){$\alpha$}
			\put(-8,43){\rotatebox{90}{$\beta$}}
			\put(97,60){\rotatebox{-90}{Accuracy (\%)}}
		\end{overpic}
		\caption{Fist step grid search results for each one of the evaluated dictionaries of sizes 50 (upper-left), 100 (upper-right), 150 (bottom-left) and 200 (bottom-right).}
	\end{figure}
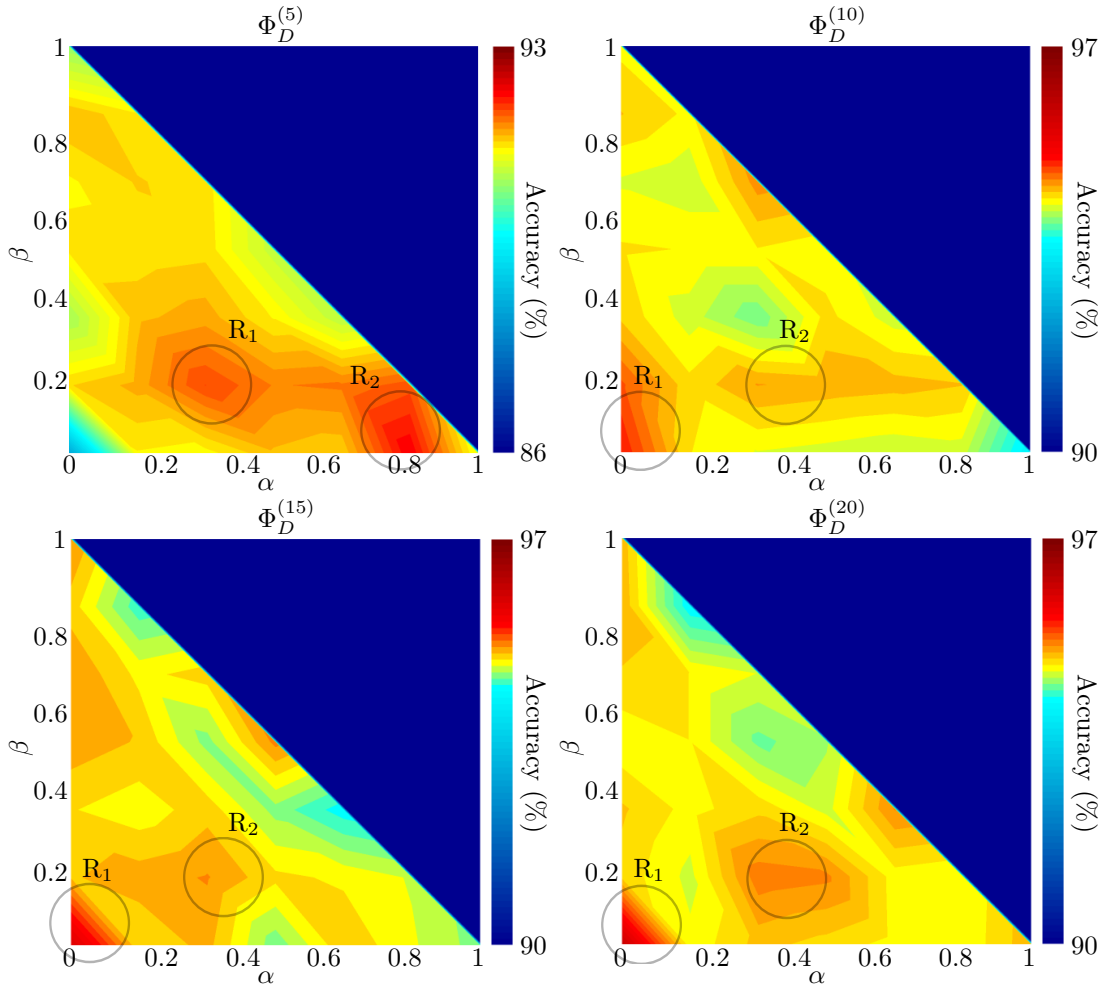
	
	We also analyzed the overall performance (taken over 10 rounds) of the classifier for each one of the evaluated dictionaries. As it can be seen in \tablename{ 3}, $\Phi_D^{(20)}$ outperforms all the others yielding the maximum (Max) recognition rate. Also, it can be seen that small structured dictionary sizes entail low classification rates. This may be due to the fact that low dimensional sparse vectors are not capable of capturing relevant information for signal classification. Otherwise, if the dimension of such vectors increases (from 100 to 200) then significant improvements are observed.
	
	\begin{table}[h!]
		\begin{center} \small
			\caption{Mean, standard deviation and maximum recognition rates obtained for each one of the evaluated dictionaries.}
			\begin{tabular}{*{4}{c}}
				\hline
				Dictionary & Classifier & Acc (\%) & Max ($\%$)\\
				\hline
				\hline
				$\Phi_D^{(5)}$ & MLP-50-50-10 & 91.25 ($\pm$0.67) & 92.23\\
				$\Phi_D^{(10)}$ & MLP-100-100-10 & 94.36 ($\pm$0.27) & 95.03\\
				$\Phi_D^{(15)}$ & MLP-150-150-10 & 94.74 ($\pm$0.27) & 95.90\\
				$\Phi_D^{(20)}$ & MLP-200-200-10 & 94.87 ($\mathbf{\pm}$0.33) & \textbf{96.20}\\
				\hline
			\end{tabular}
		\end{center}
		\label{tab: opt_pair_of_para}
	\end{table}
	
\begin{figure}[t!]
		\centering
		\begin{overpic}[width=5in]{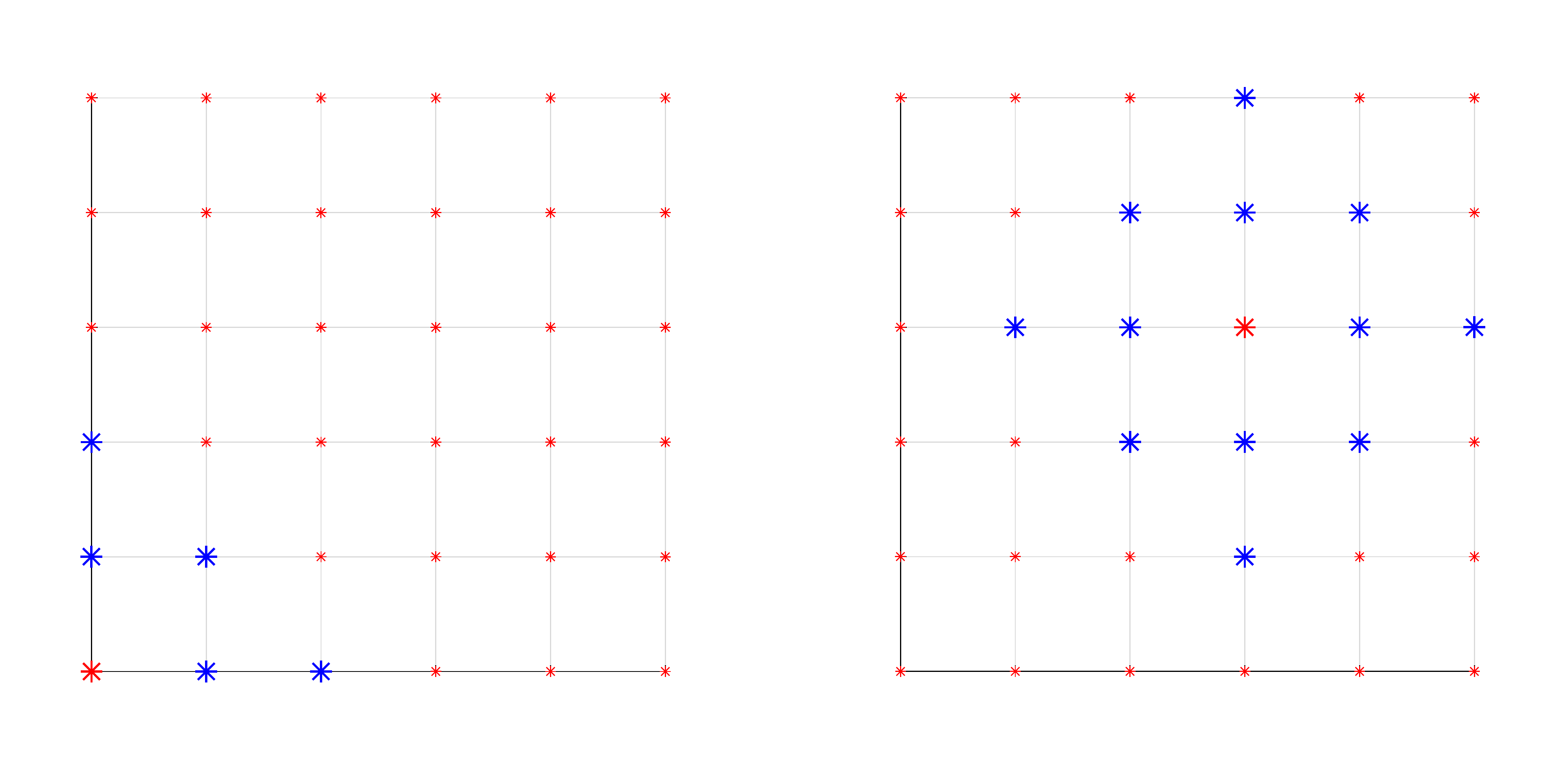}
			%\begin{overpic}[width=5in,grid,tics=5]{images/GS_FT_R1_R2.pdf}
			\put(5,2.5){\small 0}
			\put(11,2.5){\small 0.01}
			\put(18.3,2.5){\small 0.02}
			\put(25.6,2.5){\small 0.03}
			\put(32.9,2.5){\small 0.04}
			\put(40.2,2.5){\small 0.05}
			\put(55.3,2.5){\small 0.30}
			\put(62.6,2.5){\small 0.31}
			\put(69.9,2.5){\small 0.32}
			\put(77.2,2.5){\small 0.33}
			\put(84.5,2.5){\small 0.34}
			\put(92,2.5){\small 0.35}
			\put(23.5,0.5){\small $\alpha$}
			\put(75,0.5){\small $\alpha$}
			\put(-3,23){\small \rotatebox{90}{$\beta$}}
			\put(49,23){\small \rotatebox{90}{$\beta$}}
			\put(0,12.2){\small 0.01}
			\put(0,19.5){\small 0.02}
			\put(0,26.8){\small 0.03}
			\put(0,34.1){\small 0.04}
			\put(0,41.4){\small 0.05}
			\put(52,5){\small 0.14}
			\put(52,12.2){\small 0.15}
			\put(52,19.5){\small 0.16}
			\put(52,26.8){\small 0.17}
			\put(52,34.1){\small 0.18}
			\put(52,41.4){\small 0.19}
			\put(6.5,6.5){\scriptsize  \textbf{96.27}}
			\put(13.7,6.5){\scriptsize  95.93}
			\put(21,6.5){\scriptsize  95.71}
			\put(6.5,13.8){\scriptsize  96.02}
			\put(13.7,13.8){\scriptsize  95.42}
			\put(6.5,21.1){\scriptsize  95.51}
			\put(80,13.8){\scriptsize  93.69}
			\put(72.7,21.1){\scriptsize  94.52}
			\put(80,21.1){\scriptsize  \textbf{94.53}}
			\put(87.3,21.1){\scriptsize  94.38}
			\put(65.4,28.4){\scriptsize  93.56}
			\put(72.7,28.4){\scriptsize  93.64}
			\put(80,28.4){\scriptsize  94.12}
			\put(87.3,28.4){\scriptsize  94.16}
			\put(94.6,28.4){\scriptsize  93.77}
			\put(72.7,35.7){\scriptsize  94.35}
			\put(80,35.7){\scriptsize  94.01}
			\put(87.3,35.7){\scriptsize  94.07}
			\put(80,43){\scriptsize  93.84}
		\end{overpic}
		\caption{A second stage grid search taken into account regions $\textrm{R}_1$ (left) and $\textrm{R}_2$ (right) of $\Phi_D^{(20)}$.}
		\label{fig: GS_FT}
	\end{figure}
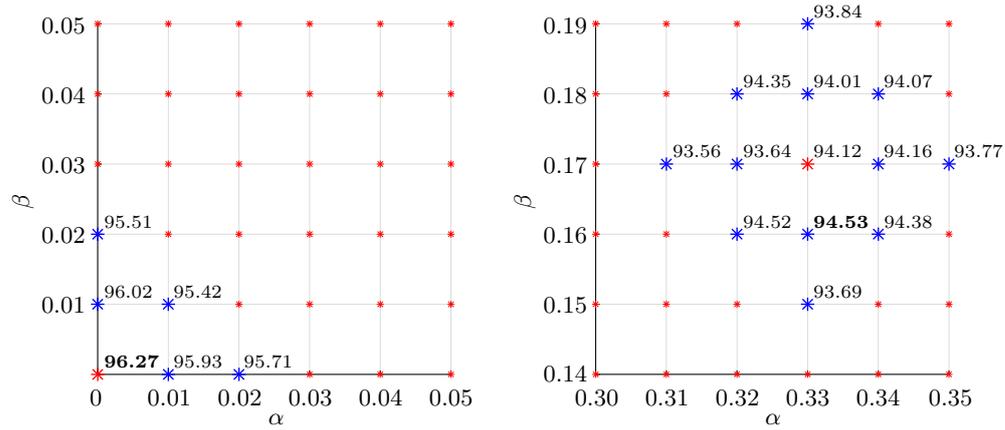

	The second stage grid search method was successfully applied to each one of the tested structured dictionaries. Results have shown that, in this case, no improvements in the recognition rates were found. Thus, the optimal pair of parameters $\alpha^*$ and $\beta^*$ are the ones found in the first stage. \figurename{ \ref{fig: GS_FT}} shows the results obtained by applying the refined grid search to regions $\textrm{R}_1$ (left) and $\textrm{R}_2$ (right) corresponding to the structured dictionary $\Phi_D^{(20)}$. It can be clearly seen that the values of $\alpha=0$ and $\beta=0$ suggest that the most discriminative atoms of a particular dictionary $\Phi$ are not only those more frequently used for signal representation, but also the ones that minimize the total signal representation error. This imply that using only the third term of the proposed combined measure, we ensure finding the most discriminative atoms of a given dictionary.
	
\end{appendices}

\end{document}